%% file: main.tex
\documentclass[10pt,journal,compsoc]{IEEEtran}

\ifCLASSOPTIONcompsoc
  \usepackage[nocompress]{cite}
\else
  \usepackage{cite}
\fi
\ifCLASSINFOpdf
\else
\fi
\pdfoutput=1

\usepackage{graphicx}
\usepackage{times}
\usepackage[utf8]{inputenc} 
\usepackage[T1]{fontenc}    
\usepackage{url}            
\usepackage{booktabs}       
\usepackage{amsfonts}       
\usepackage{nicefrac}       
\usepackage{microtype}      
\usepackage{epsfig}
\usepackage{float}
\usepackage{multicol}
\usepackage{multirow}
\usepackage{amssymb}
\usepackage{amsmath}
\usepackage{wrapfig}
\usepackage{xspace}
\usepackage{color}
\usepackage{colortbl}
\usepackage[dvipsnames, svgnames, x11names, table]{xcolor}
\usepackage[misc]{ifsym}
\usepackage{makecell}
\usepackage{soul}
\usepackage{bm}
\usepackage{wrapfig}
\usepackage{subcaption, overpic, textpos}
\usepackage{paralist}
\usepackage{tabularx}

\usepackage[pagebackref,breaklinks=true,colorlinks,citecolor=blue,urlcolor=blue,linkcolor=blue,bookmarks=false]{hyperref}

\usepackage{adjustbox}
\usepackage{pifont}
\usepackage{array}
\usepackage{enumitem}
\newcolumntype{C}[1]{>{\centering\arraybackslash}p{#1}} 

\def \pzo {\phantom{0}} 
\newcommand{\cmark}{\ding{52}\xspace}%
\newcommand{\xmark}{\ding{56}\xspace}%
\newcommand{\xmarkg}{\textcolor{lightgray}{\ding{56}}\xspace}%
\newcommand{\pmark}{\ding{58}\xspace}%

\definecolor{lightgray}{rgb}{0.8, 0.8, 0.8}
\definecolor{lgray}{rgb}{0.66, 0.66, 0.66}
\definecolor{whit_tab}{RGB}{255, 255, 255}
\definecolor{gray_tab}{RGB}{246, 246, 246}
\definecolor{oran_tab}{RGB}{252, 242, 237}
\definecolor{blue_tab}{RGB}{227, 240, 251}
\definecolor{lblu_tab}{RGB}{225, 235, 246}
\definecolor{orange_vitad}{RGB}{222, 131, 68}
\definecolor{blue_vitad}{RGB}{106, 153, 208}
\newcommand{\red}{\textcolor[RGB]{255, 36, 24}}
\newcommand{\deepred}{\textcolor[RGB]{176, 36, 24}}
\newcommand{\grayer}{\textcolor[RGB]{168, 168, 168}}

\definecolor{trajectory_green}{RGB}{126, 171, 85}
\definecolor{trajectory_yellow}{RGB}{245, 194, 66}

\definecolor{tab_others}{RGB}{235, 235, 235}
\definecolor{tab_ours}{RGB}{225, 235, 246}

\newcommand{\ccwtab}[1]{\cellcolor{whit_tab}{#1}}
\newcommand{\ccgtab}[1]{\cellcolor{gray_tab}{#1}}
\newcommand{\ccotab}[1]{\cellcolor{oran_tab}{#1}}
\newcommand{\ccbtab}[1]{\cellcolor{blue_tab}{#1}}

\usepackage[ruled,vlined,linesnumbered]{algorithm2e}

\usepackage[capitalize]{cleveref}
\crefname{section}{Sec.}{Secs.}
\Crefname{section}{Section}{Sections}
\Crefname{table}{Table}{Tables}
\crefname{table}{Tab.}{Tabs.}

\newlength\savewidth\newcommand\shline{\noalign{\global\savewidth\arrayrulewidth\global\arrayrulewidth 1pt}\hline\noalign{\global\arrayrulewidth\savewidth}}
\newcommand{\tablestyle}[2]{\setlength{\tabcolsep}{#1}\renewcommand{\arraystretch}{#2}\centering\footnotesize}
\renewcommand{\paragraph}[1]{\vspace{1.25mm}\noindent\textbf{#1}}

\newcommand{\ie}{i.e}
\newcommand{\eg}{e.g}
\newcommand{\Eg}{E.g}

\def\onedot{.\xspace}
\def\eg{\textit{e.g}\onedot} 
\def\Eg{\textit{E.g}\onedot}
\def\ie{\textit{i.e}\onedot}

\def\cf{\textit{c.f}\onedot}

\def\etc{\textit{etc}\onedot}
\def\vs{\textit{vs}\onedot}

\def\etal{\textit{et al}\onedot}

\usepackage{tablefootnote}
\usepackage{diagbox}

\makeatother

\def\irmb{iRMB}
\def\iirmb{i$^2$RMB}

\begin{document}
%

\title{EMOv2: Pushing 5M Vision Model Frontier} 

%
%
%
%

\author{Jiangning Zhang, 
        Teng Hu, 
        Haoyang He, 
        Zhucun Xue, \\ 
        Yabiao Wang, 
        Chengjie Wang, 
        Yong Liu, 
        Xiangtai Li, 
        Dacheng Tao
\IEEEcompsocitemizethanks{
\IEEEcompsocthanksitem J.~Zhang, Y.~Wang, and C.~Wang are with Youtu Lab, Tencent, China. 
\IEEEcompsocthanksitem T.~Hu is with Shanghai Jiao Tong University, Shanghai, China.
\IEEEcompsocthanksitem H.~He, Z.~Xue, and Y.~Liu are with the Institute of Cyber-Systems and Control, Zhejiang University, Hangzhou, China. 
\IEEEcompsocthanksitem X.~Li and D.~Tao are with the Nanyang Technological University, Singapore.
}
}

%
%

\markboth{IEEE TRANSACTIONS ON PATTERN ANALYSIS AND MACHINE INTELLIGENCE}
{Shell \MakeLowercase{\textit{et al.}}: Bare Advanced Demo of IEEEtran.cls for IEEE Computer Society Journals}
%



\input{secs/0_abstract.tex}

\maketitle

\IEEEdisplaynontitleabstractindextext

%
\IEEEpeerreviewmaketitle

\input{secs/1_introduction}
\input{secs/2_related_work}
\input{secs/3_method}
\input{secs/4_experiments}
\input{secs/5_conclusion}

\ifCLASSOPTIONcaptionsoff
  \newpage
\fi



{
\bibliographystyle{IEEEtran}
\bibliography{IEEEabrv,main}
}

\input{secs/6_appendix}
\end{document}

%% file: secs/0_abstract.tex
\IEEEtitleabstractindextext{
\begin{abstract}

This work focuses on developing parameter-efficient and lightweight models for dense predictions while trading off parameters, FLOPs, and performance.
Our goal is to set up the new frontier of the 5M magnitude lightweight model on various downstream tasks. 
Inverted Residual Block (IRB) serves as the infrastructure for lightweight CNNs, but no counterparts have been recognized by attention-based design. 
Our work rethinks the lightweight infrastructure of efficient IRB and practical components in Transformer from a unified perspective, extending CNN-based IRB to attention-based models and abstracting a one-residual Meta Mobile Block (MMBlock) for lightweight model design. 
Following neat but effective design criterion, we deduce a modern \textbf{I}mproved \textbf{I}nverted \textbf{R}esidual \textbf{M}obile \textbf{B}lock (\textbf{{\iirmb}}) and improve a hierarchical Efficient MOdel (\textbf{EMOv2}) with no elaborate complex structures. 
Considering the imperceptible latency for mobile users when downloading models under 4G/5G bandwidth and ensuring model performance, we investigate the performance upper limit of lightweight models with a magnitude of 5M.
Extensive experiments on various vision recognition, dense prediction, and image generation tasks demonstrate the superiority of our EMOv2 over state-of-the-art methods, \eg, EMOv2-1M/2M/5M achieve 72.3, 75.8, and 79.4 Top-1 that surpass equal-order CNN-/Attention-based models significantly. 
At the same time, EMOv2-5M equipped RetinaNet achieves 41.5 mAP for object detection tasks that surpasses the previous EMO-5M by +2.6$\uparrow$. 
When employing the more robust training recipe, our EMOv2-5M eventually achieves 82.9 Top-1 accuracy, which elevates the performance of 5M magnitude models to a new level.
Code is available at \url{https://github.com/zhangzjn/EMOv2}. 
\end{abstract}

}

%% file: secs/1_introduction.tex
\section{Introduction} \label{section:intro} 
Lightweight models are particularly crucial in resource-constrained scenarios, drawing many research efforts~\cite{efficientformerv2,edgenext,tinysam,zhou2023edgesam,xu2024rap,sfnet,lu2023rtmo} in various fields.
Early work primarily can be divided into two categories: 1) models with fewer FLOPs and faster hardware-specific inference speeds~\cite{mnetv1,mnetv2,mnetv3,ghostnet,efficientnet}, which do not emphasize parameter counts and 
perform poorly in high-resolution downstream tasks; 2) models that balance FLOPs and performance under limited parameter counts~\cite{edgenext,emo}, resulting in more compact models. With the development of computational devices, most current models achieve throughput of several thousand and latency within real-time 20ms~\cite{mvitv2,efficientformerv2,edgenext}, where computational power is not the bottleneck for small model applications, even if we strive to reduce their computational requirements. 
Additionally, edge applications iterate models rapidly, as seen in short video platforms like TikTok, where effects frequently update lightweight real-time detection algorithms and small-scale generation models. Considering the imperceptible delay in downloading models under 4G/5G bandwidth and ensuring model performance, a lightweight model of 5M magnitude is recommended as an appropriate size~\cite{speed1,speed2}. Therefore, this paper explores the upper limits of lightweight model performance with a fixed parameter count, using a 5M lightweight model as a typical representative. 

MobileNetv2~\cite{mnetv2} introduces an efficient \emph{Inverted Residual Block} (IRB) based on \emph{Depth-Wise Separable Convolution} (DW-Conv), which is widely regarded as the foundation of efficient models~\cite{efficientnet,mnetv3,mvitv1}. However, constrained by the natural induction bias of static convolution operations, the accuracy of CNN-based lightweight models is suboptimal due to the lack of global modeling capabilities. \textit{This motivates us to explore the construction of a stronger fundamental block that surpasses the IRB by introducing global modeling capabilities. }
On the other hand, benefiting from the dynamically global modeling capability of Multi-Head Self-Attention (MHSA), Vision Transformer (ViT)~\cite{vit} and its derivatives~\cite{pvtv1,pvtv2,swinv1,swinv2,zhang2021analogous,eatformer,li2023transformer,involution} have achieved significant improvements over CNNs. Some works attempt to address the quadratic computational complexity of MHSA by designing variants with linear complexity~\cite{reformer,performer}, reducing the spatial resolution of features~\cite{cvt,pvtv1,nextvit}, rearranging channels~\cite{delight}, and employing local window attention~\cite{swinv1,swinv2}, among other strategies. Recently, researchers have introduced MHSA into certain layers of lightweight CNN models to improve complex blocks~\cite{mvitv1,mvitv2,mvitv3,mobileformer,efficientformer,edgenext} or have used multiple hybrid blocks. However, such designs lack uniformity, require meticulous design, and pose higher demands for adaptation to mobile device deployment. 
So far, no works explore MHSA-based counterparts as IRB, and this inspires us to think: \emph{can we build a lightweight IRB-like infrastructure for attention-based models with only basic operators?} 

\begin{figure}[thp]
    \centering
    \includegraphics[width=1.00\linewidth]{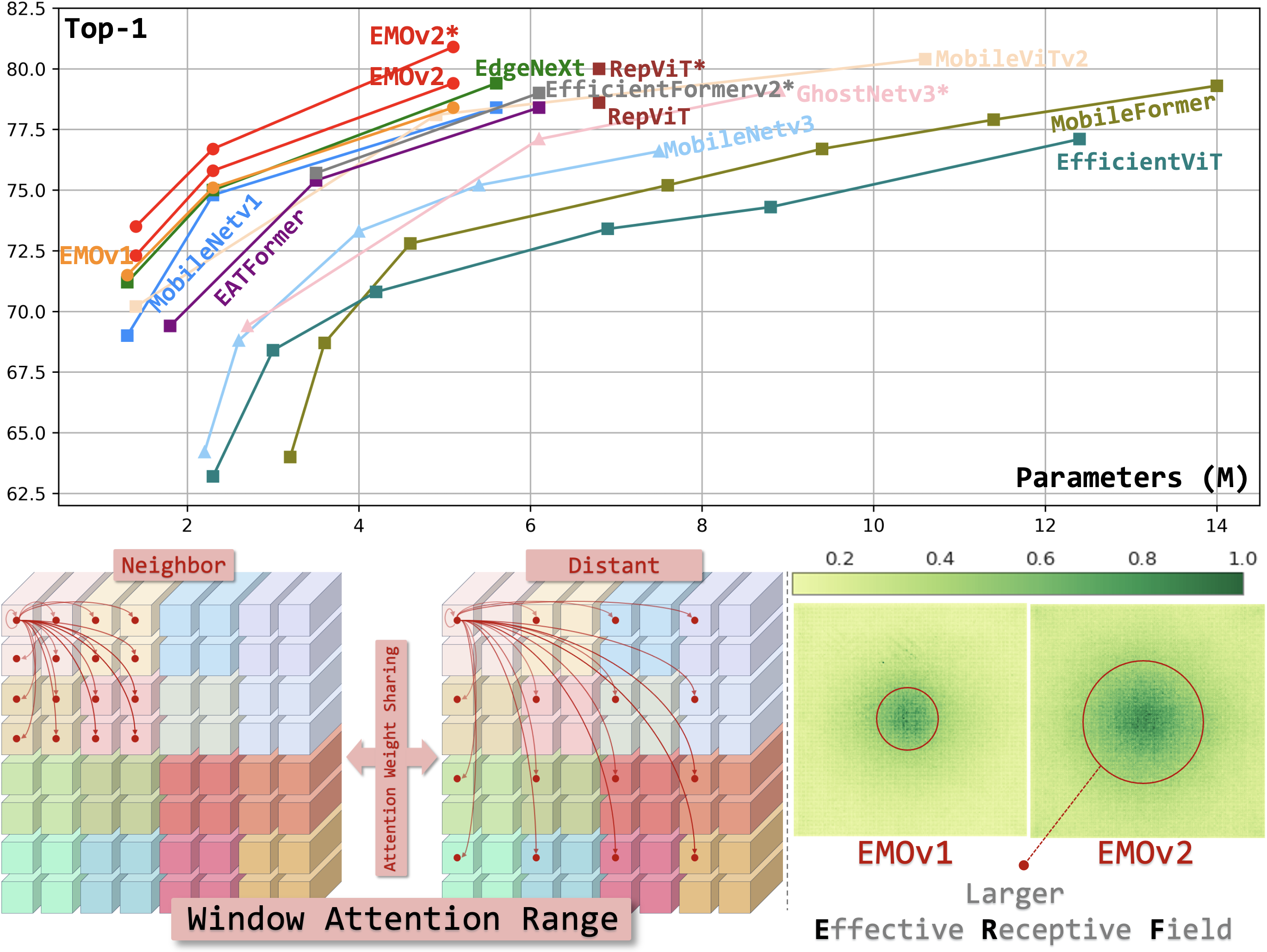}
    \caption{
    \textbf{Top:} \textit{Performance} \vs \textit{Parameters} with concurrent methods. Our EMOv2 achieves significant accuracy with fewer parameters. Superscript $\ast$: The comparison methods employ more robust training strategies described in their papers, while ours uses the strategy mentioned in \cref{table:ablation_all}(e). 
    \textbf{Bottom:} The range of token interactions varies with different window attention mechanisms. Our EMOv2, with parameter-shared spanning attention in \cref{section:iirmb}, has a larger and correspondingly stronger Effective Receptive Field (ERF).
    }
    \vspace{-1.5em}
    \label{fig:teaser}
\end{figure}

Based on the motivation above, we rethink the efficient IRB in MobileNetv2~\cite{mnetv2} and the MHSA / FFN modules in Transformer~\cite{transformer} from a unified perspective, expecting to integrate their advantages at the infrastructure design level. 
As shown in \cref{fig:emo2}-Left, while working to bring one-residual IRB with inductive bias into the attention model, we observe that MHSA/FFN submodules in two-residual Transformer share a similar meta-structure to IRB. 
Thus, we inductively abstract a one-residual Meta Mobile Block (MMBlock in \cref{section:mmb}) that takes parametric arguments' \emph{expansion ratio} \deepred{\emph{\bm{$\lambda$}}} and \emph{efficient operator} \deepred{\bm{$\mathcal{F}$}} to instantiate different modules, \ie, IRB, MHSA, and FFN. 
MMBlock reveals the consistent essence expression of the above three modules and can be regarded as an improved lightweight concentrated aggregate of Transformer. 
Furthermore, a neat yet effective \emph{Inverted Residual Mobile Block} (iRMB) is deduced that only contains fundamental Depth-Wise Convolution and the improved EW-MHSA (\cf, Sec.~\ref{section:irmb}). And we build a ResNet-like 4-phase Efficient MOdel (EMOv1) with only {\irmb}s (\cf, Sec.~\ref{section:emo1}). 

Even though EMOv1~\cite{emo} achieves promising results, it is limited by window attention that can only model the interaction of neighbor information within a local window, as shown in \cref{fig:teaser}-Bottom. This modeling approach leads to suboptimal performance in high-resolution downstream tasks due to the lack of distant information interaction. 
For instance, RetinaNet~\cite{retinanet} using EMOv1-5M only achieves 38.9 mAP that does not even reach 40. 
Recently, MobileViT~\cite{mvitv1} attempts to model long-range attention but performs moderately due to the loss of local dynamic modeling capability and a significant increase in FLOPs with higher resolutions. 
Thus, more balanced efforts between long-range modeling and lower GFlops are needed.
To overcome these challenges, we explore the procedure of attention computation and discover that the neighbor window attention map can be reused to model the correlation between distant positions. 
Based on this, we design a novel spanning mechanism~\cref{section:emo2} (\ie, SEW-MHSA) that simultaneously models neighbor and distant features. 
As shown in \cref{fig:teaser}, this mechanism does not increase the number of parameters and only adds a small number of FLOPs. It significantly enhances the model's effective receptive field, thereby improving performance in high-resolution downstream tasks (\cref{section:exp_downstream}).
Additionally, we improve the detailed structure of {\iirmb} to enhance the performance further and explore different training strategies to maximize the model's potential in mainstream image classification tasks. 
Detailed comparison with state-of-the-art methods can be viewed in \cref{fig:teaser}. 
Due to the neat structural design, {\iirmb} can be easily extended to various downstream tasks, achieving significant and consistent performance improvements. 
Specifically, we apply EMOv2 to the temporal dimension for video recognition, and V-EMO-v2 obtains 65.2 Top-1 accuracy with 5.9M parameters on Kinetics-400 for video classification that surpasses UniFormer-XXS's 63.2 with 9.8M parameters. 
In addition, we enhance the recently popular UNet and DiT architectures for image segmentation and generation across multiple downstream tasks based on this module (\cref{section:variant}). \Eg, U-EMO-v2 obtains 88.3mAcc with 21.3M parameters on HRF; D-EMO-v2 achieves 46.3/9.6 FID in generating 256$\times$256 ImageNet images with 400K training steps on S/XL scales, which significantly surpasses DiT's 68.4/19.5. 

In summary, we make the following significant extensions over the preliminary conference version (EMO~\cite{emo} at ICCV'23): 
\begin{enumerate}
\item Based on the abstracted one-residual \textit{Meta Mobile Block} for lightweight model design, we extend the {\irmb} to a powerful {\iirmb} block. Specifically, we design a parameter-sharing spanning attention mechanism, enabling interaction between neighborhood and distant spatial features within a single module without increasing the model's parameter count. This mechanism is also compatible with EW-MHSA, achieving efficient feature modeling for mobile applications. Additionally, we improve the post-attention and large local kernel structures to further enhance model performance.
\item We construct a 4-stage EMOv2 backbone solely based on the deduced {\iirmb} block. This model significantly improves performance while maintaining the similar parameter count as EMOv1. For instance, EMOv2-5M achieves a +1.0$\uparrow$ improvement over EMOv1-5M in classification tasks. The performance gap widens further in high-resolution downstream tasks, with improvements of +1.7$\uparrow$ and +2.6$\uparrow$ mAP using SSDLite and RetinaNet, respectively. We also explore the impact of stronger training strategies on model performance, validating the model's scaling capability, with EMOv2-5M reaching up to 82.9 Top-1 accuracy. 
\item Thanks to the general, neat, and powerful design of {\iirmb}, we can easily extend it to a series of tasks, constructing various lightweight versions of different types of structures and achieving significant improvements. Finally, we provide detailed studies and experimental analysis to build our attention-based lightweight models in \cref{section:exp_ablation}.
\item We re-write the entire draft and add a more comprehensive discussion on close related works. We open-source our EMOv2 for the community. 

\end{enumerate}

%% file: secs/2_related_work.tex
\section{Related Work} 
\label{section:related}
\vspace{1mm} \noindent\textbf{Lightweight CNN Models.}
With the increasing demands of neural networks for mobile vision applications, efficient model design has attracted extensive attention from researchers in recent years. SqueezeNet~\cite{squeezenet} replaces 3x3 filters with 1x1 filters and decreases channel numbers to reduce model parameters, while Inceptionv3~\cite{inceptionv3} factorizes the standard convolution into asymmetric convolutions. Later, MobileNet~\cite{mnetv1} introduces depth-wise separable convolution to alleviate a large amount of computation and parameters, followed in subsequent lightweight models~\cite{mnetv2,ghostnet,Li2022SFNetFA,sfnet}. Besides the above hand-craft methods, researchers exploit automatic architecture design in the pre-defined search space~\cite{mnetv3,efficientnet,efficientformerv2}. Specifically, RepViT~\cite{repvit} leverages the re-parameterization technique to enhance model performance, while recent GhostNetV3~\cite{ghostnetv3} has further incorporated a Knowledge Distillation (KD) strategy. MobileNetv4~\cite{mnetv4} employs both NAS algorithm and KD strategy to achieve impressive results, where a strong training recipe has already become a trend in lightweight model research. 
We draw on lightweight design principles from the CNN domain, such as depth-wise convolution and inverted residual designs, and integrate them with attention mechanisms to construct a stronger hybrid module. 

\vspace{1mm} \noindent\textbf{Hugging Vision Transformer with CNN.} Since ViT~\cite{vit} first introduces Transformer structure~\cite{transformer} into visual tasks, massive improvements have successfully been developed. DeiT~\cite{deit} provides a benchmark for efficient transformer training, subsequent works~\cite{pvtv1,swinv1} employ ResNet-like~\cite{resnet} pyramid structure to form pure Transformer-based models for dense prediction tasks. However, the absence of 2D convolution will potentially increase the optimization difficulty and damage the model accuracy for lacking local inductive bias, so researchers~\cite{hassanin2022visual,islam2022recent} concentrate on how to better integrate convolution into Transformer for obtaining stronger hybrid models. \Eg, work~\cite{ceit} incorporates convolution design into FFN, works~\cite{cpvt,uniformer} regard convolution as the positional embedding for enhancing inductive bias of the model, and works~\cite{cvt} for attention and QKV calculations, respectively. Recently, MogaNet~\cite{moganet} encapsulates conceptual convolutions and gated aggregation into a compact module, and SHViT~\cite{shvit} uses a depthwise convolution layer for local feature aggregation or conditional position embedding. 
However, the above methods are still confined to the MetaFormer~\cite{metaformer} architecture, where each block contains two residual connections. 
EMOv1 studies how to build a neat but effective lightweight model based on an improved one-residual attention block. 
In contrast, this paper further investigates the parameter-sharing mechanism for window attention, enabling it to simultaneously model neighbor and distant information interactions, thereby significantly enhancing the performance of downstream tasks.

\vspace{1mm} \noindent\textbf{Effective Transformer Improvements.}
Researchers~\cite{edgenext,lightvit} have started to lighten Transformer-based models for low computational power. Tao~\etal~\cite{lightvit} introduces additional learnable tokens to capture global dependencies efficiently, and Chen~\etal~\cite{lightvit} design a parallel structure of MobileNet and Transformer with a two-way bridge in between. Works~\cite{rest,edgevit} improve an efficient Transformer block by borrowing convolution operation, while EdgeNeXt~\cite{edgenext} absorbs effective Res2Net~\cite{res2net} and transposed channel attention~\cite{xcit}. MobileVit series~\cite{mvitv1,mvitv2,mvitv3} fuse improved MobileViT blocks with Mobile blocks~\cite{mnetv2}. Recent EfficientFormerV2~\cite{efficientformerv2} uses the NAS algorithm to search hardware-friendly modules, while ViG~\cite{vig} introduces a gating mechanism to facilitate the interaction of sequential and spatial information. However, most current approaches require \textit{elaborate complex modules}, which limits the mobility and usability of the model. How to balance parameters, computation, and accuracy while designing easy-to-use lightweight models still needs further exploration. 

\vspace{1mm} \noindent\textbf{RNN-reinvented Models.} 
Due to the quadratic growth in computational complexity of Transformers with the number of tokens, some RNN-based models~\cite{he2024pointrwkv,vrwkv,mambarwkv} have gradually gained attention, with Mamba~\cite{mamba} and RWKV~\cite{rwkv} being the primary representatives. Zhu~\etal~\cite{vmamba} proposes vision Mamba, which applies SSM to visual tasks, while Duan~\etal~\cite{vrwkv} also introduces a vision version based on RWKV. Recently, works~\cite{efficientvmamba,mobilemamba} explore the application of Mamba in lightweight visual tasks. 
These methods can seamlessly integrate into our proposed Meta Mobile Block, yielding favorable results. However, considering the verified stable performance of transformers across various fields, this paper explores improvements to the attention module based on a windowed operation. 

%% file: secs/3_method.tex
\section{Methodology} \label{section:method}

\begin{figure*}[htp]
    \centering
    \includegraphics[width=1.0\linewidth]{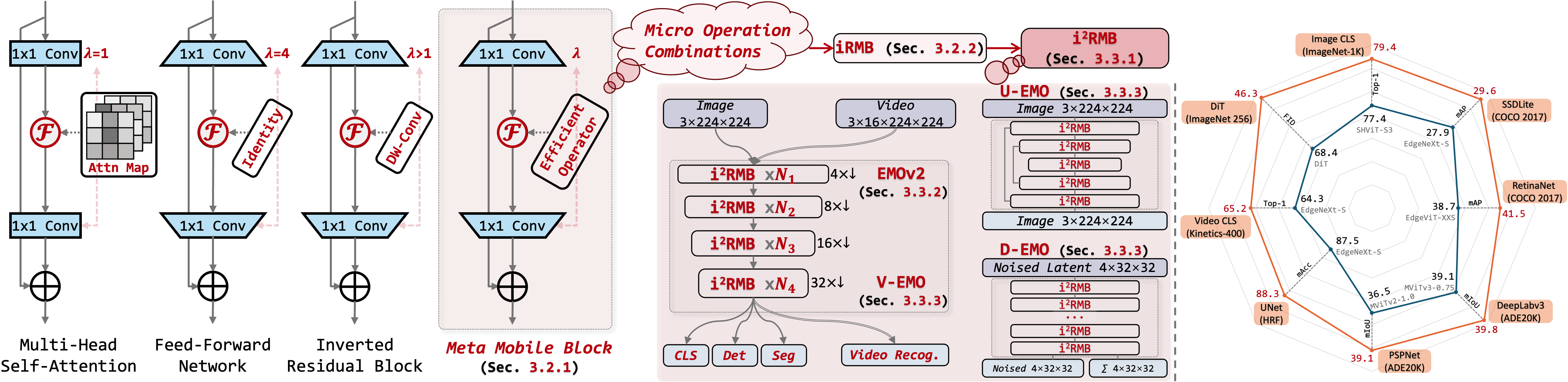}
    \caption{
    \textbf{Left}: Abstracted unified \deepred{\textbf{\emph{Meta-Mobile Block}}} from \emph{Multi-Head Self-Attention}, \emph{Feed-Forward Network}~\cite{transformer}, and \emph{Inverted Residual Block}~\cite{mnetv2} (\cf Sec~\ref{section:mmb}). The inductive block can be deduced into specific modules using different \emph{expansion ratio} \deepred{\emph{\bm{$\lambda$}}} and \emph{efficient operator} \deepred{\bm{$\mathcal{F}$}}. 
    \textbf{Middle}: We construct a family of vision models based on our {{\iirmb}} module: 4-stage \textit{EMOv2}, composed solely of the deduced \emph{{\iirmb}} (\cf Sec~\ref{section:irmb}), for various perception tasks (image classification, detection, and segmentation in \cref{section:exp_downstream}). Additionally, we introduce the temporally extended \textit{V-EMO} for video classification, the U-EMO based on an encoder-decoder architecture, and D-EMO to replace the Transformer block in DiT~\cite{dit}. These downstream models are typically built based on the \iirmb.
    \textbf{Right}: Performance comparison with different SoTAs on various tasks. 
    }
    \label{fig:emo2}
\end{figure*}

\subsection{Criteria for General Lightweight Model} 
\label{section:cri}

When designing light-weight visual models for mobile usages, we advocate the following criteria subjectively and empirically that an efficient model should satisfy as much as possible: 
\textbf{\ding{192} Usability.} Neat implementation that does not use complex operators and is easy to optimize for applications. 
\textbf{\ding{193} Uniformity.} As few core modules as possible to reduce model complexity and accelerate deployment. 
\textbf{\ding{194} Efficiency and Effectiveness.} Balancing parameters and calculations with accuracy trade-off. 
\textbf{\ding{195} Generalization.} Easily applied to perception tasks such as classification, detection, and segmentation, as well as to generative tasks, while compatible with architectures like ResNet and U-Net. 
We make a summary of current efficient models in \cref{table:model_criterion}: 
\emph{1)} Performance of MobileNet series~\cite{mnetv1,mnetv2,mvitv3} is now seen to be slightly lower, and its parameters are slightly higher than counterparts. 
\emph{2)} Recent MobileViT series~\cite{mvitv1,mvitv2,mvitv3} achieve notable performances, but they suffer from higher FLOPs and slightly complex modules. 
\emph{3)} EdgeNeXt~\cite{edgenext} and EdgeViT~\cite{edgevit} obtain pretty results, but their basic blocks also consist of elaborate modules. 
\emph{4)} RepViT~\cite{repvit} employs multiple fundamental modules and introduces a re-parameterization strategy, while EfficientFormerV2~\cite{efficientformerv2} utilizes NAS to search for hardware-friendly models, and EfficientVMamba~\cite{efficientvmamba} introduces a new SSM module. 
\emph{5)} MogaNet~\cite{moganet} achieves a balance between performance and efficiency without introducing new complex operators.
Comparably, the design principle of our EMO/v2 follows the above criteria without introducing complicated operations (\cf, \cref{section:emo2}) while still obtaining impressive results on multiple vision tasks (\cf, \cref{section:exp}). 
Additionally, EMOv2 can be easily transferred to other models for various tasks, such as video classification, UNet-based image segmentation, and diffusion-based image generation (\cf, \cref{section:emo2}). 

\begin{table}[t]
    \centering
    \caption{\textbf{Criterion comparison for current efficient models}. \textbf{\ding{192}}: Usability; \textbf{\ding{193}}: Uniformity; \textbf{\ding{194}}: Efficiency and Effectiveness; \textbf{\ding{195}}: Generalization. \cmark: Satisfied. \pmark: Partially satisfied. \xmark: Unsatisfied.}
    \label{table:model_criterion}
    \renewcommand{\arraystretch}{1.0}
    \setlength\tabcolsep{6.0pt}
    \resizebox{1.0\linewidth}{!}{
        \begin{tabular}{p{4.6cm}<{\raggedright} p{1.0cm}<{\centering} p{1.0cm}<{\centering} p{1.0cm}<{\centering} p{1.0cm}<{\centering}}
            \toprule[0.17em]
            Method \vs Criterion                            & \ding{192}    & \ding{193}    & \ding{194}    & \ding{195}    \\
            \hline
            MobileNet Series~\cite{mnetv1,mnetv2,mvitv3}    & \cmark        & \cmark        & \pmark        & \xmark        \\
            MobileViT Series~\cite{mvitv1,mvitv2,mvitv3}    & \pmark        & \pmark        & \pmark        & \xmark        \\
            EdgeNeXt~\cite{edgenext}                        & \pmark        & \xmark        & \cmark        & \xmark        \\
            EdgeViT~\cite{edgevit}                          & \cmark        & \pmark        & \pmark        & \xmark        \\
            RepViT~\cite{repvit}                            & \cmark        & \xmark        & \cmark        & \xmark        \\
            EfficientFormerV2~\cite{efficientformerv2}      & \cmark        & \pmark        & \cmark        & \xmark        \\
            EfficientVMamba~\cite{efficientvmamba}          & \xmark        & \xmark        & \pmark        & \xmark        \\
            MogaNet~\cite{moganet}                          & \cmark        & \cmark        & \cmark        & \xmark        \\
            \hline
            EMOv1                                             & \cmark        & \cmark        & \cmark        & \xmark        \\
            EMOv2                                           & \cmark        & \cmark        & \cmark        & \cmark        \\
            \toprule[0.12em]
        \end{tabular}
    }
\end{table}

\subsection{Efficient MOdel (EMOv1)} \label{section:emo1}
\subsubsection{Meta Mobile Block} \label{section:mmb}
\vspace{1mm} \noindent\textbf{Motivation.} 1) Recent Transformer-based works~\cite{focal,swinv1,cswin,inception,gmlp,mlpmixer,resmlp} are dedicated to improving spatial token mixing under the MetaFormer~\cite{metaformer} for high-performance network. CNN-based \textit{Inverted Residual Block}~\cite{mnetv2} (IRB) is recognized as the infrastructure of efficient models~\cite{mnetv2,efficientnet}, but little work has been done to explore attention-based counterpart. This inspires us to build a lightweight IRB-like infrastructure for attention-based models. 2) While working to bring one-residual IRB with inductive bias into the attention model, we stumble upon two underlying sub-modules (\ie, FFN and MHSA) in two-residual Transformer that happen to share a similar structure to IRB. This inspires us to integrate these elements into a unified block representation, thereby constructing a more shallow foundational visual backbone. Compared to each ViT block, which contains two residual connections, our approach simplifies the architecture. 

\vspace{1mm} \noindent\textbf{Induction.} 
We rethink Inverted Residual Block in MobileNetv2~\cite{mnetv2} with core MHSA and FFN modules in Transformer~\cite{transformer}, and inductively abstract a general Meta Mobile Block (MMBlock) in \cref{fig:emo2}, which takes parametric arguments \emph{expansion ratio} \deepred{\emph{\bm{$\lambda$}}} and \emph{efficient operator} \deepred{\bm{$\mathcal{F}$}} to instantiate different modules. We argue that \textit{the MMBlock can reveal the consistent essence expression of the above three modules, and MMBlock can be regarded as an improved lightweight concentrated aggregate of Transformer}. Also, this is the basic motivation for our elegant and easy-to-use EMO/v2, which only contains one deduced {{\irmb}}/{{\iirmb}} absorbing advantages of lightweight CNN and Transformer. Taking image input $\boldsymbol{X} (\in\mathbb{R}^{C \times H \times W})$ as an example, MMBlock firstly use an expansion $\text{MLP}_{e}$ with output/input ratio equaling \deepred{\emph{\bm{$\lambda$}}} to expand channel dimension:

\vspace{-0.3em}
\begin{equation}
    \begin{aligned}
        \boldsymbol{X}_{e} = \text{MLP}_{e}(\boldsymbol{X}) (\in\mathbb{R}^{\deepred{\emph{\bm{$\lambda$}}}C \times H \times W}) \text{.}
    \end{aligned}
\end{equation}
Then, intermediate operator $\deepred{\bm{\mathcal{F}}}$ enhance image features further, \eg, identity operator, static convolution, dynamic MHSA, \etc. Considering that MMBlock is suitable for efficient network design, we present $\deepred{\bm{\mathcal{F}}}$ as the concept of \emph{efficient operator}, formulated as: 

\vspace{-0.3em}
\begin{equation}
    \begin{aligned}
        \boldsymbol{X}_{f} = \deepred{\bm{\mathcal{F}}}(\boldsymbol{X}_{e}) (\in\mathbb{R}^{\deepred{\emph{\bm{$\lambda$}}}C \times H \times W}) \text{.}
    \end{aligned}
\end{equation}
Finally, a shrinkage $\text{MLP}_{s}$ with inverted input/output ratio equaling \deepred{\emph{\bm{$\lambda$}}} to shrink channel dimension:
\begin{equation}
    \begin{aligned}
        \boldsymbol{X}_{s} = \text{MLP}_{s}(\boldsymbol{X}_{f}) (\in\mathbb{R}^{C \times H \times W}) \text{,}
    \end{aligned}
\end{equation}
where a residual connection is used to get the final output $\boldsymbol{Y} = \boldsymbol{X} + \boldsymbol{X}_{s} (\in\mathbb{R}^{C \times H \times W})$. 
For clarity, notice that normalization and activation functions are omitted.

\begin{figure}[thp]
    \centering
    \includegraphics[width=0.80\linewidth]{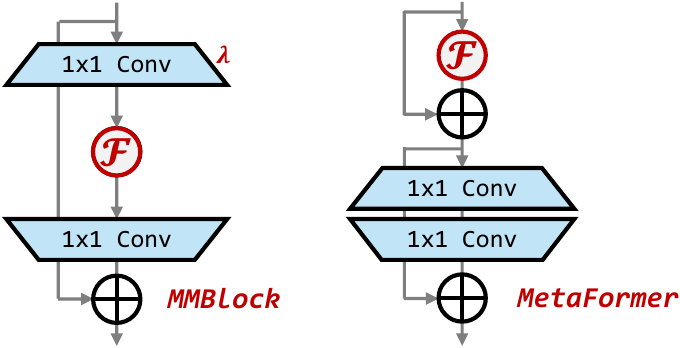}
    \caption{Meta-paradigm comparison between our MMBlock and MetaFormer~\cite{metaformer}. We integrate $\deepred{\bm{\mathcal{F}}}$ into expended FFN to construct a more streamlined and shallower single-module block.}
    \label{fig:com_meta}
\end{figure}

\vspace{1mm} \noindent\textbf{Relation to MetaFormer.}
We reveal the differences between our \emph{Meta Mobile Block} and \emph{MetaFormer}~\cite{metaformer} in \cref{fig:com_meta}. 
\textbf{\emph{1)}} From the structure, two-residual MetaFormer contains two sub-modules with two skip connections, while our Meta Mobile Block contains only one sub-module that covers one-residual IRB in the field of lightweight CNN. Also, shallower depths require less memory access and save costs~\cite{shufflenetv2} that is more general and hardware-friendly for optimization. 
\textbf{\emph{2)}} From the motivation, MetaFormer is the induction of high-performance Transformer/MLP-like models, while our Meta Mobile Block is the induction of efficient IRB in MobileNetv2~\cite{mnetv2} and effective MHSA/FFN in Transformer~\cite{transformer,vit} for designing lightweight infrastructure.
\textbf{\emph{3)}} Inductive one-residual Meta Mobile Block can be regarded as a conceptual extension of two-residual MetaFormer in the lightweight field. We hope our work inspires more future research dedicated to lightweight model design domain based on attention.
\textbf{\emph{4)}} From the result, our instantiated EMOv2-5M (w/ 5.1M \#Params and 1.0G FLOPs) exceeds instantiated PoolFormer-S12 (w/ 11.9M \#Params and 1.8G FLOPs) by +2.1$\uparrow$, illustrating that a stronger efficient operator makes a advantage. We further replace Token Mixer in MetaFormer with $\deepred{\bm{\mathcal{F}}}$ in {{\irmb}} and build a 5.3M model. Compared with EMOv1-5M, it only achieves 77.5 Top-1 on ImageNet-1k that is -0.9$\downarrow$ than our model, meaning that our proposed Meta Mobile Block has a better advantage for constructing lightweight models than two-residual MetaFormer.

\begin{table}[tp]
    \centering
    \caption{Complexity and Maximum Path Length analysis of modules. Input/output feature maps are in $\mathbb{R}^{C \times W \times W}$, $L = W^2$, $l = w^2$, $W$ and $w$ are feature map size and window size, while $k$ and $G$ are kernel size and group number.}
    \label{table:mpl}
    \renewcommand{\arraystretch}{1.0}
    \resizebox{1.0\linewidth}{!}{
        \begin{tabular}{p{1.5cm}<{\raggedleft} p{2.1cm}<{\raggedright} p{2.9cm}<{\raggedright} p{2.2cm}<{\raggedright}}
        \toprule[0.17em]
        Module & \#Params & FLOPs & MPL \\
        \hline
        \rowcolor{whit_tab} MHSA   & $4(C+1)C$       & $8C^2L+4CL^2+3L^2$  & $O(1)$    \\
        \rowcolor{whit_tab} W-MHSA   & $4(C+1)C$       & $8C^2L+4CLl+3Ll$  & $O(Inf)$    \\
        \hline
        \rowcolor{whit_tab} Conv  & $(Ck^{2}/G+1)C$ & $(2Ck^{2}/G)LC$     & $O(2W/(k-1))$  \\
        \rowcolor{whit_tab} DW-Conv  & $(k^{2}+1)C$ & $(2k^{2})LC$     & $O(2W/(k-1))$  \\
        \toprule[0.12em]
        \end{tabular}
    }
    \vspace{-1.0em}
\end{table}

\subsubsection{Micro Designs for Deducted {{\irmb}}} \label{section:irmb}
\begin{figure}[tp]
    \centering
    \includegraphics[width=1.0\linewidth]{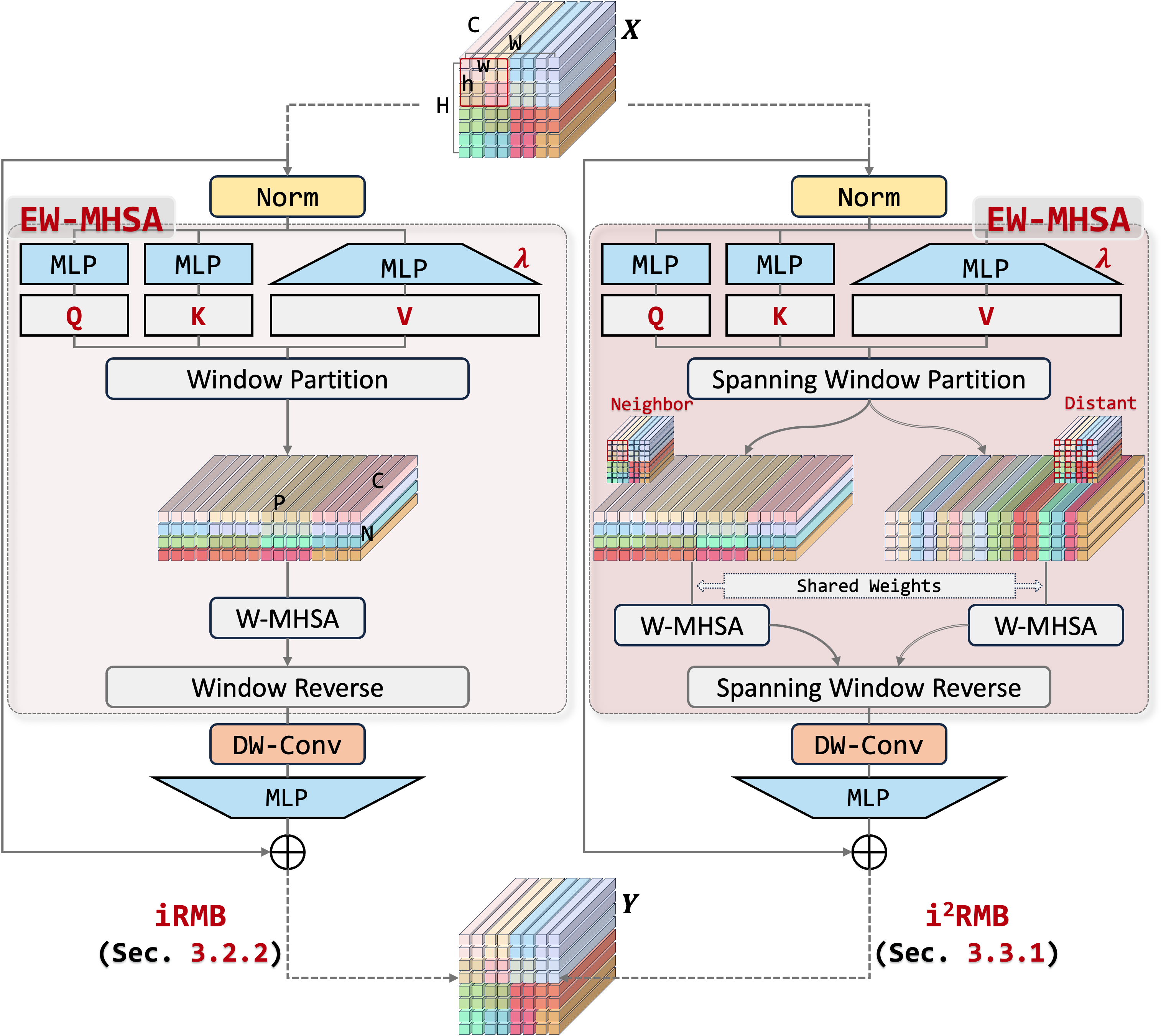}
    \caption{Detailed implementation comparison of the Inverted Residual Mobile Block (\textit{{\irmb}} in \cref{section:irmb}) and the improved version (\textit{{\iirmb}} in \cref{section:iirmb}). {\iirmb} designs a parameter-sharing spanning window attention mechanism that simultaneously models the interaction of distant and close window information.} 
    \label{fig:iirmb}
\end{figure}

Based on the inductive Meta Mobile Block, we instantiate an effective modern \emph{Inverted Residual Mobile Block} ({{\irmb}}) for lightweight architecture design from a microscopic view in \cref{fig:iirmb}. 

\vspace{1mm} \noindent\textbf{Design principle.} Following criteria in \cref{section:cri}, $\deepred{\bm{\mathcal{F}}}$ in iRMB is modeled as cascaded \emph{MHSA} and \emph{Convolution} operations, formulated as $\deepred{\bm{\mathcal{F}}}(\cdot) = \text{Conv}(\text{MHSA}(\cdot))$. This design absorbs CNN-like efficiency to model local features and Transformer-like dynamic modeling capability to learn long-distance interactions. However, naive implementation can lead to unaffordable expenses for two main reasons: 
\noindent\textit{1)} \deepred{\emph{\bm{$\lambda$}}} is generally greater than one that the intermediate dimension would be multiple to input dimension, causing quadratic \deepred{\emph{\bm{$\lambda$}}} increasing of parameters and computations. Therefore, components of $\deepred{\bm{\mathcal{F}}}$ should be independent or linearly dependent on the number of channels. 
\noindent\textit{2)} FLOPs of MHSA is proportional to the quadratic of total image pixels, so the cost of a naive Transformer is unaffordable for downstream application. The specific influences can be seen in \cref{table:mpl}.

\vspace{1mm} \noindent\textbf{Expanded Window MHSA.} Parameters and FLOPs for obtaining $Q$,$K$ in Window MHSA (W-MHSA)~\cite{swinv1} is quadratic of the channel. 
Given the input $\boldsymbol{X}$ $(\in\mathbb{R}^{C \times H \times W})$, we obtain channel-unexpanded $Q$ and $K$ $(\in\mathbb{R}^{C \times H \times W})$ to compute the attention matrix $M$ more efficiently, while the expanded $V$ $(\in\mathbb{R}^{\deepred{\emph{\bm{$\lambda$}}}C \times H \times W})$ is used to capture finer-grained visual features. The essence of this expanding mechanism is that $M$ models only the spatial positional relationships and is independent of the number of channels in $V$. This improvement is termed EW-MHSA, which is more applicable. Specifically, Window Partition operation flattens each feature map $F \in \{Q, K, V\}$ into $N$ non-overlapping patches with each sequence length $P$=$w \times h$, where $N$=$H \times W / P$. The corresponding dimensional transformation can be described by the following formula: $[B, C, H, W] \rightarrow [BHW/P, C, P]$, and vice versa for the Window Reverse operation. To put it more directly, $w$=$4$, $h$=$4$, $P$=$16$, and $N$=$4$ for example in ~\cref{fig:iirmb}. 

\vspace{1mm} \noindent\textbf{Structural deduction.} Combining lightweight Depth-Wise Convolution (DW-Conv) and efficient EW-MHSA to trade-off model cost and accuracy, the process of the designed {{\irmb}} can be formulated as follows: 

\begin{equation}
    \begin{aligned}
        \deepred{\bm{\mathcal{F}}}(\cdot) = \text{DW-Conv}(\text{EW-MHSA}(\cdot)) \text{.}
    \end{aligned}
\end{equation}
This cascading manner can increase the expansion speed of the receptive field and reduce the maximum path length of the model to $O(2W/(k-1+2w))$, which has been experimentally verified with consistency in \cref{section:exp_ablation}.

\vspace{1mm} \noindent\textbf{Flexibility.} Empirically, current transformer-based methods~\cite{edgenext,uniformer,moat,efficientformerv2,moganet} reach a consensus that inductive CNN in shallow layers while global Transformer in deep layers composition could benefit the performance. Unlike recent EdgeNeXt that employs different blocks for different depths, our iRMB satisfies the above design principle using only two switches to control whether two modules are used (Code level is also concise in \#Supp). Therefore, we can easily implement the use of EW-MHSA for more semantic modeling only in the deeper layers, \ie, stage-3 and stage-4.

\vspace{1mm} \noindent\textbf{Efficient equivalent implementation.} 
MHSA is typically employed in channel-consistent projection (\deepred{\emph{\bm{$\lambda$}}}=1), indicating that the FLOPs of multiplying the attention matrix by the expanded $\boldsymbol{X}_{e}$ (\deepred{\emph{\bm{$\lambda$}}}>1) will increase by a factor of \deepred{\emph{\bm{$\lambda$}}} - 1. Fortunately, the information flow from $\boldsymbol{X}$ to the expanded $V$ ($\boldsymbol{X}_{e}$) involves only linear operations, allowing us to derive an equivalent proposition: "\emph{When the number of groups in $\text{MLP}_{e}$ equals the number of heads in $\text{EW-MHSA}$, the result of the multiplication remains unchanged when the order is exchanged}." To reduce FLOPs, matrix multiplication before $\text{MLP}_{e}$ is used by default, referred to as pre-attention. 

\vspace{1mm} \noindent\textbf{Boosting naive transformer.} To assess iRMB performance, we set \deepred{\emph{\bm{$\lambda$}}} to 4 and replace standard Transformer structure in columnar DeiT~\cite{deit} and pyramidal PVT~\cite{pvtv1}. As shown in \cref{table:toy_irmb}, we surprisingly found that {{\irmb}} can improve performance with fewer parameters and computations in the same training setting, especially for the columnar ViT. And the newly proposed {{\iirmb}} further boosts the performance significantly. This proves that the one-residual {{\irmb}}/{{\iirmb}} has obvious advantages over the two-residual Transformer in the lightweight model. 

\begin{table}[tp]
    \centering
    \caption{Toy experiments for assessing {{\irmb}} and {{\iirmb}}.}
    \label{table:toy_irmb}
    \renewcommand{\arraystretch}{1.2}
    \resizebox{1.0\linewidth}{!}{
        \begin{tabular}{p{3.0cm}<{\raggedright} p{1.5cm}<{\raggedright} p{1.5cm}<{\raggedright} p{1.5cm}<{\raggedright}}
        \toprule[0.17em]
        Model & \#Params $\downarrow$ & FLOPs $\downarrow$ & Top-1 $\uparrow$  \\
        \hline
        DeiT-Tiny~\cite{deit}   & \pzo5.7M                                           & 1.3G                                          & 72.2  \\
        DeiT-Tiny w~/~{{\irmb}}        & \pzo4.9M & 1.1G & 74.3\scriptsize{\red{${~\text{+}2.1\%\uparrow}$}}  \\
        DeiT-Tiny w~/~{{\iirmb}}        & \pzo5.0M & 1.3G & 75.0\scriptsize{\red{${~\text{+}2.8\%\uparrow}$}}  \\
        \hline
        PVT-Tiny~\cite{pvtv1}   & 13.2M                                          & 1.9G                                          & 75.1  \\
        PVT-Tiny w~/~{{\irmb}}         & 11.7M    & 1.8G     & 75.4\scriptsize{\red{${~\text{+}0.3\%\uparrow}$}}  \\
        PVT-Tiny w~/~{{\iirmb}}         & 11.9M    & 1.9G     & 76.1\scriptsize{\red{${~\text{+}1.0\%\uparrow}$}}  \\
        \toprule[0.12em]
        \end{tabular}
    }
    \vspace{-1.0em}
\end{table}

\vspace{1mm} \noindent\textbf{Parallel design of $\deepred{\bm{\mathcal{F}}}$.} We also implement the parallel structure of DW-Conv and EW-MHSA with half the number of channels in each component, and some configuration details are adaptively modified to ensure the same magnitude. Comparably, this parallel model gets 78.1 (-0.3$\downarrow$) Top-1 in ImageNet-1k dataset with 5.1M parameters and 964M FLOPs (+63M$\uparrow$ than EMOv1-5M), but its throughput will slow down by about -7\%$\downarrow$. 
\begin{center}\vspace{-.2em}
\tablestyle{6pt}{1.05}
\setlength\tabcolsep{6.0pt}
\begin{tabular}{ccccc}
Manner & \#Params. & FLOPs & Top1 & Throughput \\
\shline
Parallel & 5.1M & 964M & 78.1 & 1618.4 \\
Cascaded (Ours) & 5.1M & 903M & 78.3 & 1731.7 \\
\end{tabular}\vspace{-.2em}
\end{center}
This phenomenon is also discussed in the work~\cite{shufflenetv2} that: "Network fragmentation reduces the degree of parallelism". 

\subsection{Parameter-Efficient Extension (EMOv2)}
Even though EMOv1 achieves satisfactory results, it only models the interaction of neighbor information within a local window, which has a limited effective receptive field (ERF) (see \cref{fig:teaser}). This limitation leads to suboptimal performance in high-resolution downstream tasks. We further explore the performance frontier of lightweight models based on this module with a negligible increase in model parameters. Specifically, we leverage the principles of attention computation to reuse the neighbor window attention map for uniform sampling over a global window size, resulting in a novel spanning module termed SEW-MHSA. This mechanism simultaneously models both neighbor and distant features without increasing the number of parameters. Additionally, we elaborately improve structural details to further enhance the model's performance. 

\subsubsection{Improved Inverted Residual Mobile Block ({{\iirmb}})} \label{section:iirmb} 
To avoid a significant increase in the number of parameters, we optimize the EW-MHSA and DW-Conv modules to construct a more powerful {{\iirmb}} module in \cref{fig:iirmb}.

\vspace{1mm} \noindent\textbf{Spanning attention for EW-MHSA.} This paper explores the potential of lightweight models under limited parameters, \ie, mainly 5M for most mobile scenarios. We observe that in EW-MHSA, the attention map only computes feature interactions within windows. While this alleviates the computational explosion of global attention, it inevitably reduces the flow of the receptive field. Therefore, we extend the computation of the attention map to a parallel fusion of neighbor and distant window attention, introducing \textit{Spanning Window Partition and Reverse} steps to achieve this goal. 
Compared to the naive Window Partition described in \cref{section:irmb}, this operation involves two parallel window partitions that separately segment the shared $Q$, $K$, and $V$ into neighbor and distant partitions. In the former, each window contains only adjacent features. In the latter, feature selection within the window is performed based on a stride of $[H/h, W/w]$. 
This allows for feature interaction at different distances simultaneously, and its transformation can be described by the following formula: $[B, C, H, W] \rightarrow \{ [BHW/P, C, P]_{neibor}, [BHW/P, C, P]_{distant} \}$. Followed by two parameter-shared MHSA, this powerful improvement is termed SEW-MHSA. 
The computation of Q and K remains in the non-extended dimension, following {{\irmb}}. This approach has two benefits: 1) A single module can accommodate global information in one forward pass, which is advantageous for downstream tasks requiring high resolution. 2) The parallel operation does not introduce additional parameters, reusing the parameters and computations of K, Q, and V, and only adds an extra attention map computation, thereby enhancing model accuracy with minimal computational cost. 

\vspace{1mm} \noindent\textbf{Non-linearity for post-attention.} We introduce a nonlinear activation function in the V computation of the attention mechanism, further filtering features before multiplying them with the attention map. This differs from the pre-attention described in \cref{section:irmb}, referred to as post-attention, which improves model performance without increasing the number of parameters. 

\vspace{1mm} \noindent\textbf{Large kernel for local modeling.} {{\irmb}} uses a kernel size of 3 for the DW-Conv in local modeling. Smaller values limit the model's receptive field. {{\iirmb}} further investigates the impact of large kernels on accuracy. Considering the depth-wise modeling approach, this does not significantly increase the number of model parameters. Additionally, this structure provides the model with positional information, allowing it to achieve downstream structures without additional position embedding design.

\vspace{1mm} \noindent\textbf{Structural deduction.} Combining lightweight Depth-Wise Convolution (DW-Conv) and efficient EW-MHSA to trade-off model cost and accuracy, the process of the designed {{\irmb}} can be formulated:
\begin{equation}
    \begin{aligned}
        \deepred{\bm{\mathcal{F}}}(\cdot) = \text{DW-Conv}(\text{SEW-MHSA}(\cdot)) \text{.}
    \end{aligned}
\end{equation}

\vspace{1mm} \noindent\textbf{Accessibility analysis.} 
Due to the fact that {{\iirmb}} only includes convolution and multi-head self-attention operators, the constructed EMOv2 is built by stacking identical standard modules without employing hardware-aware search structures, and it uses a serial structure without multiple branches. This design is highly compatible with hardware acceleration, potentially offering strong generalizability for different hardware platforms and applications.

\subsubsection{Macro Design of EMOv2 for Dense Prediction} \label{section:emo2} 
Based on the above criteria, we design a ResNet-like 4-phase Efficient MOdel (EMO) based on a series of {{\irmb}}s for dense applications in our previous work~\cite{emo}. 
In this extension work, we build a stronger vision backbone EMOv2 by the powerful {{\iirmb}}s, as shown in \cref{fig:emo2}-\textbf{Right}. \\
\textbf{\emph{1)}} For the overall framework, EMOv2 consists of only {{\iirmb}} without diversified modules$^{\textbf{\ding{193}}}$, which is a departure from recent efficient methods~\cite{mvitv1,edgenext} in terms of designing idea. \\
\textbf{\emph{2)}} For the specific module, {{\iirmb}} consists of only convolution and multi-head self-attention without other complex operators$^{\textbf{\ding{192}}}$. Also, benefitted by DW-Conv, {{\iirmb}} can adapt to down-sampling operation through the stride and does not require any position embeddings for introducing inductive bias to MHSA$^{\textbf{\ding{193}}}$. The comparison of the requirements for embedding across different methods is shown in \cref{table:supp_train_recipe}. \\
\textbf{\emph{3)}} For the configuration of different-scale models, we employ gradually increasing expansion rates and channel numbers, and detailed configurations are shown in \cref{table:model_variants}. Results for basic classification and downstream dense prediction tasks in \cref{section:exp} demonstrate the superiority of our {{\iirmb}} over SoTA lightweight methods on magnitudes of 1M, 2M, and core-focused 5M$^{\textbf{\ding{194}}}$. \\
\textbf{\emph{4)}} {{\iirmb}} can be easily extended to other foundational architectures and accomplish corresponding tasks$^{\textbf{\ding{195}}}$, such as temporal extension, UNet variant, and DiT-like model in \cref{section:variant}. 

\vspace{1mm} \noindent\textbf{Configuration details.} Since MHSA is better suited for modeling semantic features for deeper layers, we only turn it on at stage-3/4 following previous works~\cite{edgenext,uniformer,moat}. Note that this never violates the uniformity criterion, as the shutdown of MHSA was a special case of {{\iirmb}} structure. To further increase the stability of EMO, BN~\cite{bn}+SiLU~\cite{gelu} are bound to DW-Conv while LN~\cite{ln}+GeLU~\cite{gelu} are bound to SEW-MHSA, and {{\iirmb}} is competent for down-sampling operations.

\begin{table}[tp]
    \centering
    \caption{Core configurations of EMOv2 variants.}
    \label{table:model_variants}
    \renewcommand{\arraystretch}{1.2}
    \resizebox{1.0\linewidth}{!}{
        \begin{tabular}{cccc}
            \toprule[0.17em]
            Items           & EMOv2-1M               & EMOv2-2M               & EMOv2-5M \\
            \hline
            Depth           & [ 2, 2, 8, 3 ]            & [ 3, 3, 9, 3 ]            & [ 3, 3, 9, 3 ] \\
            Emb. Dim.       & [ 32, 48, 80, 180 ]      & [ 32, 48, 120, 200 ]      & [ 48, 72, 160, 288 ] \\
            Exp. Ratio      & [ 2.0, 2.5, 3.0, 3.5 ]    & [ 2.0, 2.5, 3.0, 3.5 ]    & [ 2.0, 3.0, 4.0, 4.0 ] \\
            \toprule[0.12em]
        \end{tabular}
    }
\end{table}

\begin{table}[htp]
    \centering
    \caption{Ablation study on components in {{\irmb}}/{{\iirmb}}.}
    \label{table:components_F}
    \renewcommand{\arraystretch}{1.0}
    \setlength\tabcolsep{3.0pt}
    \resizebox{1.0\linewidth}{!}{
        \begin{tabular}{ccl|ccl}
        \toprule[0.2em]
        \multicolumn{3}{c}{EMOv1~\cite{emo}} & \multicolumn{3}{c}{EMOv2} \\
        \cmidrule(lr){1-3} \cmidrule(lr){4-6}
        \text{EW-MHSA} & \text{DW-Conv} & \pzo\pzo\pzo Top-1 & \text{SEW-MHSA} & \text{DW-Conv} & \pzo\pzo\pzo Top-1 \\
        \hline
        \rowcolor{whit_tab} \xmarkg &   \xmarkg & \pzo73.5 & \xmarkg & \xmarkg & \pzo73.5 \\
        \rowcolor{whit_tab} \cmark  &   \xmarkg & \pzo76.6\scriptsize{\red{${~\text{+}3.1\uparrow}$}} & \cmark  &   \xmarkg & \pzo77.7\scriptsize{\red{${~\text{+}4.2\uparrow}$}} \\
        \rowcolor{whit_tab} \xmarkg &   \cmark  & \pzo77.6\scriptsize{\red{${~\text{+}4.1\uparrow}$}} & \xmarkg &   \cmark  & \pzo78.1\scriptsize{\red{${~\text{+}4.6\uparrow}$}} \\
        \rowcolor{whit_tab} \cmark  &   \cmark  & \pzo78.4\scriptsize{\red{${~\text{+}4.9\uparrow}$}} & \cmark  &   \cmark  & \pzo79.4\scriptsize{\red{${~\text{+}5.9\uparrow}$}} \\
        \toprule[0.2em]
        \end{tabular}
    }
    \vspace{-1.0em}
\end{table}

\vspace{1mm} \noindent\textbf{Importance of instantiated efficient operator.} Our defined \emph{efficient operator} \deepred{\bm{$\mathcal{F}$}} contains two core modules, \ie, (S)EW-MHSA and DW-Conv. In \cref{table:components_F}, we conduct an ablation experiment to study the effect of both modules in {{\irmb}}/{{\iirmb}}. The first row means that neither (S)EW-MHSA nor DW-Conv is used, \ie, the model is almost composed of MLP layers with several DW-Conv for down-sampling, and \deepred{\bm{$\mathcal{F}$}} degenerates to Identity operation. Surprisingly, this model still produces a respectable result, \ie, 73.5 Top-1. Comparatively, results of the second and third rows demonstrate that each component contributes to the performance, \eg, +3.1$\uparrow$ and +4.1$\uparrow$ when adding DW-Conv and EW-MHSA for EMO, respectively, while +4.2$\uparrow$ and +4.6$\uparrow$ for EMOv2. Our approach achieves the best result when both components are used. Besides, this experiment illustrates that the specific instantiation of {{\irmb}}/{{\iirmb}} is very important to model performance.

\vspace{1mm} \noindent\textbf{Order of operators.} Based on EMOv1-5M, we switch the order of DW-Conv/EW-MHSA and find a slight -0.6$\downarrow$, and a similar -0.7$\downarrow$ drop is also observed in EMOv2 when switching DW-Conv/SEW-MHSA. Therefore, (S)EW-MHSA performs first by default. 

\vspace{1mm} \noindent\textbf{Performance gains over EMOv1.} 
The improved EMOv2-5M achieves a Top-1 accuracy of 79.4, surpassing EMOv1-5M by +1.0$\uparrow$, without significantly increasing parameters and FLOPs. 
\begin{wrapfigure}{r}{3.0cm}
    \centering
    \vspace{-1.0em}
    \includegraphics[width=1.0\linewidth]{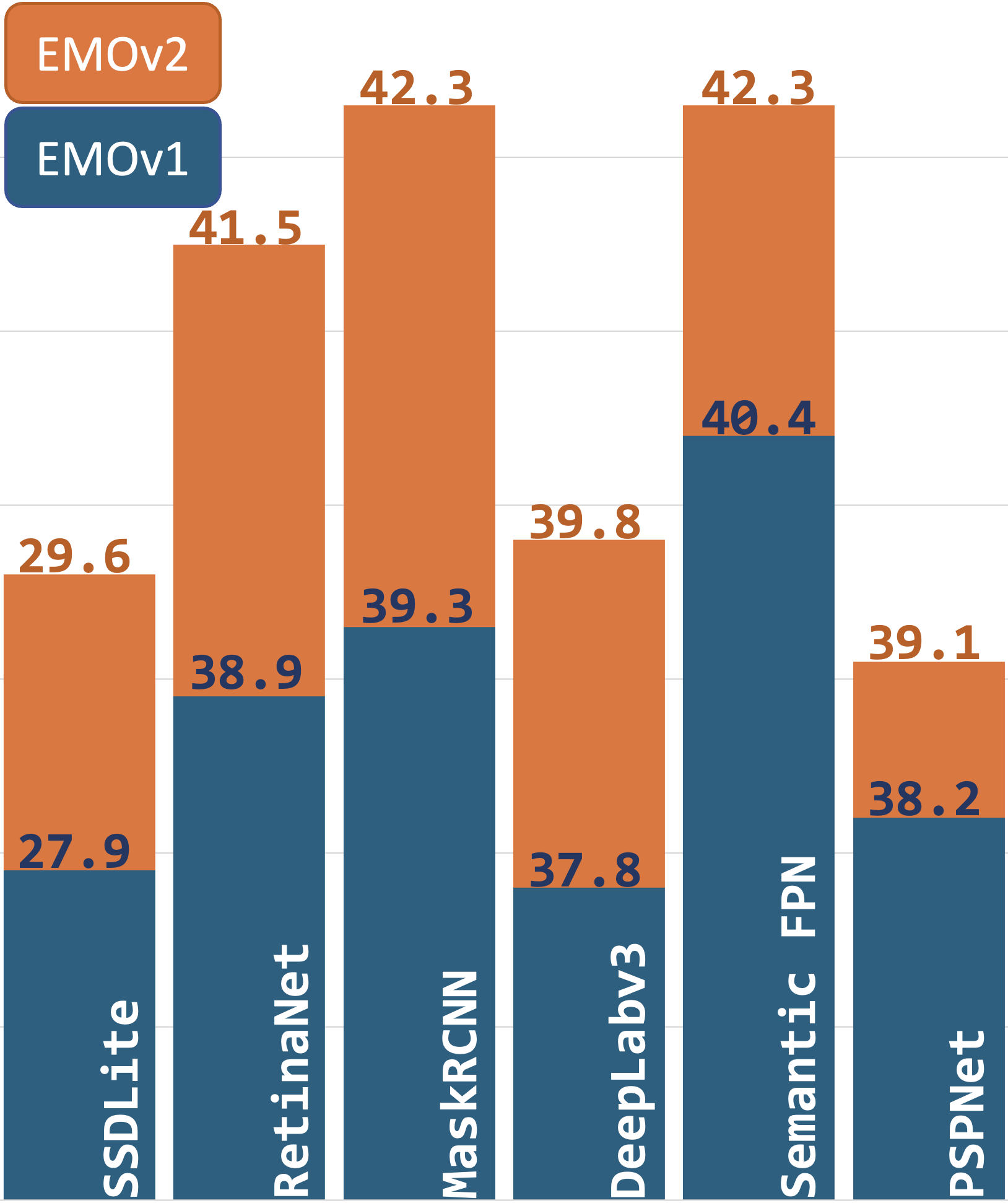}
    \caption{Downstream gains of EMOv2-5M over EMOv1-5M.}
    \label{fig:v1v2}
    \vspace{-3.0em}
\end{wrapfigure}
Additionally, it demonstrates notable improvements across various high-resolution downstream tasks. For instance, in popular detection and segmentation tasks, as shown in \cref{fig:v1v2}, EMOv2 consistently achieves an enhancement of 1$\sim$3 points across different frameworks. 

\subsubsection{{{\iirmb}}-Centric Omni-Task Transformation} \label{section:variant} 
Thanks to the general, neat, and powerful {{\iirmb}} design, we can easily extend it to various tasks in this extension work, as illustrated in \cref{fig:emo2}: 
\textbf{\textit{1)}} video classification (V-EMO) extends the {{\iirmb}} to the temporal dimension, 
\textbf{\textit{2)}} UNet-based image segmentation (U-EMO) replaces the original convolutional blocks with {{\iirmb}}, and 
\textbf{\textit{3)}} diffusion-based image generation (D-EMO) replaces naive Transformer blocks with {{\iirmb}}. 
We construct various lightweight versions of different types of structures and conduct extensive experiments to demonstrate the effectiveness and generalizability of {{\iirmb}} in \cref{section:exp_downstream}.

%% file: secs/4_experiments.tex
\section{Experimental Results}
\label{section:exp}

\subsection{Image Classification} \label{section:exp_cls}
\vspace{1mm} \noindent\textbf{Setup.} Different SoTA methods use various training recipes that could lead to potentially unfair comparisons, and we have summarized and compared these training strategies in \cref{table:supp_train_recipe}. In contrast, our training strategy is weaker, yet it achieves impressive results without employing strong training tricks. 
All experiments are conducted on the ImageNet-1K dataset~\cite{imagenet} without using additional datasets or pre-trained models. Each model is trained for a standard 300 epochs from scratch at a resolution of 224$\times$224 by default. The AdamW~\cite{adamw} optimizer is employed with betas (0.9, 0.999), a weight decay of 5e$^{-2}$, a learning rate of 6e$^{-3}$, and a batch size of 2,048. We use a Cosine scheduler~\cite{cosine} with 20 warmup epochs, Label Smoothing 0.1~\cite{label_smoothing}, stochastic depth~\cite{drop_path}, and RandAugment~\cite{randaugment} during training. However, LayerScale~\cite{ls}, Dropout~\cite{dropout}, MixUp~\cite{mixup}, CutMix~\cite{cutmix}, Random Erasing~\cite{random_erasing}, Position Embeddings~\cite{vit}, Token Labeling~\cite{token_labeling}, and Multi-Scale training~\cite{mvitv1} are \emph{\textbf{disabled}}. EMOv2 is implemented based on TIMM~\cite{timm}.

\begin{table}[tp]
    \centering
    \caption{Performance of our EMOv1/v2 with different lightweight model training recipes.}
    \label{table:sota_recipe}
    \renewcommand{\arraystretch}{1.2}
    \setlength\tabcolsep{6.0pt}
    \resizebox{1.\linewidth}{!}{
        \begin{tabular}{lccccc}
        \toprule[0.10em]
        Recipe & MNetv3~\cite{mnetv3} & DeiT~\cite{deit} & EdgeNeXt~\cite{edgenext} & Vim~\cite{vmamba} & Ours \\
        \hline
        EMOv1~\cite{emo} & \textit{NaN} & 78.1 & 78.3 & 77.9 & 78.4 \\
        EMOv2 & \textit{NaN} & 78.8 & 79.1 & 78.5 & 79.4 \\
        \toprule[0.10em]
        \end{tabular}
    }
    \vspace{-1.5em}
\end{table}

\vspace{1mm} \noindent\textbf{Results analysis.} 
We evaluate our method against SoTA models on three small magnitudes, and the quantitative results are presented in \cref{table:cls_imagenet}. 
Notably, our method achieves the best results without utilizing complex modules and strong training recipes employed by recent works, such as NAS in MobileNetv4~\cite{mnetv4} and re-parameterization in RepViT~\cite{repvit}.
For example, the smallest EMOv2-1M achieves a SoTA Top-1 accuracy of 72.3, surpassing the CNN-based MobileNetv3-L-0.50~\cite{mnetv3} by +3.5$\uparrow$ with nearly half the parameters, and the Transformer-based MobileViTv2-0.5~\cite{mvitv2} by +2.1$\uparrow$ with only 61\% of the FLOPs. 
The larger EMOv2-2M achieves a SoTA Top-1 accuracy of 75.8 with only 487M FLOPs, nearly half of MobileVit-XS~\cite{mvitv1} but with a +1.0$\uparrow$ improvement. 
Comparatively, the latest EdgeViT-XXS~\cite{edgevit} achieves a lower Top-1 accuracy of 74.4 while requiring +78\%$\uparrow$ more parameters and +14\%$\uparrow$ more FLOPs, whereas tiny-MOAT-0~\cite{moat} requires +48\%$\uparrow$ more parameters and +64\%$\uparrow$ more FLOPs to achieve a similar result. 
Consistently, EMOv2-5M demonstrates a superior trade-off between \#Params. (5.1M), FLOPs (1.0G), and accuracy (79.4), proving to be more efficient than contemporary counterparts. 
For example, it achieves +0.9$\uparrow$ over EATFormer-Tiny~\cite{eatformer} with better efficiency. 
When we further employ the KD training strategy (TResNet~\cite{tresnet} with 83.9 accuracy as the teacher model), our three-magnitude EMOv2 models achieve 73.5, 76.7, and 80.9 Top-1 accuracy, respectively. 
This represents an increase of +2.0$\uparrow$, +1.6$\uparrow$, and +2.5$\uparrow$ compared to our previous conference method~\cite{emo}. Moreover, these results significantly exceed the latest models using strong training strategies, such as RepViT~\cite{repvit}, EfficientFormerV2~\cite{efficientformerv2}, GhostNetV3~\cite{ghostnetv3}, and MobileNetv4~\cite{mnetv4}. 

\begin{table}[htp]
    \centering
    \caption{Classification performance comparison among different kinds of backbones on ImageNet-1K dataset in terms of 5M-magnitude, as well as 1M-magnitude and 2M models. \protect\sethlcolor{whit_tab}\hl{White}, \protect\sethlcolor{gray_tab}\hl{grey}, \protect\sethlcolor{oran_tab}\hl{orange}, and \protect\sethlcolor{blue_tab}\hl{blue} backgrounds indicate CNN-based, \protect\sethlcolor{gray_tab}\hl{Transformer-based}, \protect\sethlcolor{oran_tab}\hl{RNN-based}, and our \protect\sethlcolor{blue_tab}\hl{EMO series}, respectively. This kind of display continues for all subsequent experiments. \textcolor{gray}{Gray} indicates the results obtained from the original paper. Comprehensive suggested models are marked in \textbf{bold}. Unit: \#Params with (M) and FLOPs with (M). Abbreviations: MNet $\rightarrow$ MobileNet; MViT $\rightarrow$ MobileViT; MFormer $\rightarrow$ MobileFormer. $\ast$: Neural Architecture Search (NAS) for elaborate structures. $\dagger$: Using knowledge distillation. $\ddagger$: Re-parameterization strategy. 
    $\ast$: Using stronger training strategy displayed in \cref{table:ablation_all}(e). 
    } 
    \label{table:cls_imagenet}
    \renewcommand{\arraystretch}{1.23}
    \setlength\tabcolsep{5.0pt}
    \resizebox{1.\linewidth}{!}{
        \begin{tabular}{ccccccr}
        \toprule[0.17em]
        & Model & \#Params $\downarrow$ & FLOPs $\downarrow$ & Reso. & Top-1 & Venue\pzo\pzo \\
        \hline
        \multirow{9}{*}{\rotatebox{90}{\makecell[c]{\textbf{1M mMagnitude}}}}
        & \ccwtab{MNetv1-0.50~\cite{mnetv1}}               & \ccwtab{1.3}  & \ccwtab{149}  & \ccwtab{$224^{2}$} & \ccwtab{63.7}  & \ccwtab{arXiv'1704} \\
        & \ccwtab{MNetv3-L-0.50~\cite{mnetv3}}             & \ccwtab{2.6}  & \ccwtab{69}   & \ccwtab{$224^{2}$} & \ccwtab{68.8}  & \ccwtab{ICCV'19}  \\
        & \ccgtab{MViTv1-XXS~\cite{mvitv1}}                & \ccgtab{1.3}  & \ccgtab{364}  & \ccgtab{$256^{2}$} & \ccgtab{69.0}  & \ccgtab{ICLR'22}  \\
        & \ccgtab{MViTv2-0.5~\cite{mvitv2}}                & \ccgtab{1.4}  & \ccgtab{466}  & \ccgtab{$256^{2}$} & \ccgtab{70.2}  & \ccgtab{arXiv'22} \\
        & \ccgtab{EdgeNeXt-XXS~\cite{edgenext}}            & \ccgtab{1.3}  & \ccgtab{261}  & \ccgtab{$256^{2}$} & \ccgtab{71.2}  & \ccgtab{ECCVW'22} \\
        & \ccgtab{EATFormer-Mobile~\cite{eatformer}}            & \ccgtab{1.8}  & \ccgtab{360}  & \ccgtab{$224^{2}$} & \ccgtab{69.4}  & \ccgtab{IJCV'24} \\
        & \ccbtab{\pzo\pzo~\ding{73}~EMOv1-1M~\cite{emo}}                         & \ccbtab{1.3}  & \ccbtab{261}  & \ccbtab{$224^{2}$} & \ccbtab{71.5}  & \ccbtab{ICCV'23}  \\
        & \ccbtab{\ding{72}~\textbf{EMOv2-1M\pzo}}                        & \ccbtab{1.4}  & \ccbtab{285}  & \ccbtab{$224^{2}$} & \ccbtab{\textbf{72.3}}  & \ccbtab{-} \\
        & \ccbtab{\ding{72}~\textbf{EMOv2-1M}}$\dagger$                        & \ccbtab{1.4}  & \ccbtab{285}  & \ccbtab{$224^{2}$} & \ccbtab{\textbf{73.5}}  & \ccbtab{-} \\
        \hline
        \multirow{18}{*}{\rotatebox{90}{\makecell[c]{\textbf{2M Magnitude}}}}
        & \ccwtab{MNetv2-1.40~\cite{mnetv2}}               & \ccwtab{6.9}  & \ccwtab{585}  & \ccwtab{$224^{2}$} & \ccwtab{74.7}  & \ccwtab{CVPR'18}  \\
        & \ccwtab{MNetv3-L-0.75~\cite{mnetv3}}             & \ccwtab{4.0}  & \ccwtab{155}  & \ccwtab{$224^{2}$} & \ccwtab{73.3}  & \ccwtab{ICCV'19}  \\
        & \ccwtab{FasterNet-T0~\cite{fasternet}}             & \ccwtab{3.9}  & \ccwtab{\textcolor{gray}{340}}   & \ccwtab{$224^{2}$} & \ccwtab{71.9}  & \ccwtab{CVPR'23}  \\
        & \ccwtab{GhostNetV3-0.5x~\cite{ghostnetv3}}$\dagger,\ddagger$             & \ccwtab{2.7}  & \ccwtab{\textcolor{gray}{48}}   & \ccwtab{$224^{2}$} & \ccwtab{69.4}  & \ccwtab{arXiv'2404}  \\
        & \ccwtab{MNetv4-Conv-S~\cite{mnetv4}}$\ast\dagger$             & \ccwtab{3.8}  & \ccwtab{\textcolor{gray}{200}}   & \ccwtab{$224^{2}$} & \ccwtab{73.8}  & \ccwtab{arXiv'2404}  \\
        & \ccgtab{MoCoViT-1.0~\cite{mocovit}}              & \ccgtab{5.3}  & \ccgtab{147}  & \ccgtab{$224^{2}$} & \ccgtab{74.5} & \ccgtab{arXiv'22}  \\
        & \ccgtab{PVTv2-B0~\cite{pvtv2}}                   & \ccgtab{3.7}  & \ccgtab{572}  & \ccgtab{$224^{2}$} & \ccgtab{70.5} & \ccgtab{CVM'22}    \\
        & \ccgtab{MViTv1-XS~\cite{mvitv1}}                 & \ccgtab{2.3}  & \ccgtab{986}  & \ccgtab{$256^{2}$} & \ccgtab{74.8} & \ccgtab{ICLR'22}   \\
        & \ccgtab{MFormer-96M~\cite{mobileformer}}         & \ccgtab{4.6}  & \ccgtab{96}   & \ccgtab{$224^{2}$} & \ccgtab{72.8} & \ccgtab{CVPR'22}   \\
        & \ccgtab{EdgeNeXt-XS~\cite{edgenext}}             & \ccgtab{2.3}  & \ccgtab{538}  & \ccgtab{$256^{2}$} & \ccgtab{75.0} & \ccgtab{ECCVW'22}  \\
        & \ccgtab{EdgeViT-XXS~\cite{edgevit}}              & \ccgtab{4.1}  & \ccgtab{557}  & \ccgtab{$256^{2}$} & \ccgtab{74.4} & \ccgtab{ECCV'22}   \\
        & \ccgtab{tiny-MOAT-0~\cite{moat}}                 & \ccgtab{3.4}  & \ccgtab{800}  & \ccgtab{$224^{2}$} & \ccgtab{75.5} & \ccgtab{ICLR'23}   \\
        & \ccgtab{EfficientViT-M1~\cite{evit}}             & \ccgtab{3.0}  & \ccgtab{\textcolor{gray}{167}}  & \ccgtab{$224^{2}$} & \ccgtab{68.4} & \ccgtab{CVPR'23}   \\
        & \ccgtab{EfficientFormerV2-S0~\cite{efficientformerv2}}$\ast\dagger$             & \ccgtab{3.5}  & \ccgtab{\textcolor{gray}{400}}  & \ccgtab{$224^{2}$} & \ccgtab{75.7} & \ccgtab{ICCV'23}   \\
        & \ccgtab{EATFormer-Lite~\cite{eatformer}}            & \ccgtab{3.5}  & \ccgtab{910}  & \ccgtab{$224^{2}$} & \ccgtab{75.4}  & \ccgtab{IJCV'24} \\
        & \ccbtab{\pzo\pzo~\ding{73}~EMOv1-2M~\cite{emo}}                         & \ccbtab{2.3}  & \ccbtab{439}  & \ccbtab{$224^{2}$} & \ccbtab{75.1} & \ccbtab{ICCV'23}   \\
        & \ccbtab{\ding{72}~\textbf{EMOv2-2M\pzo}}                        & \ccbtab{2.3}  & \ccbtab{487}  & \ccbtab{$224^{2}$} & \ccbtab{\textbf{75.8}} & \ccbtab{-} \\
        & \ccbtab{\ding{72}~\textbf{EMOv2-2M}}$\dagger$                        & \ccbtab{2.3}  & \ccbtab{487}  & \ccbtab{$224^{2}$} & \ccbtab{\textbf{76.7}} & \ccbtab{-} \\
        \hline
        \multirow{32}{*}{\rotatebox{90}{\makecell[c]{\textbf{5M Magnitude}}}}
        & \ccwtab{MNetv3-L-1.25~\cite{mnetv3}}             & \ccwtab{7.5}  & \ccwtab{356}  & \ccwtab{$224^{2}$} & \ccwtab{76.6} & \ccwtab{ICCV'19}   \\
        & \ccwtab{EfficientNet-B0~\cite{efficientnet}}     & \ccwtab{5.3}  & \ccwtab{399}  & \ccwtab{$224^{2}$} & \ccwtab{77.1} & \ccwtab{ICML'19}   \\
        & \ccwtab{FasterNet-T2~\cite{fasternet}}             & \ccwtab{15.0}  & \ccwtab{\textcolor{gray}{1910}}   & \ccwtab{$224^{2}$} & \ccwtab{78.9}  & \ccwtab{CVPR'23}  \\
        & \ccwtab{RepViT~\cite{repvit}}$\ddagger$             & \ccwtab{6.8}  & \ccwtab{\textcolor{gray}{1100}}  & \ccwtab{$224^{2}$} & \ccwtab{78.6} & \ccwtab{CVPR'24}   \\
        & \ccwtab{RepViT~\cite{repvit}}$\dagger,\ddagger$             & \ccwtab{6.8}  & \ccwtab{\textcolor{gray}{1100}}  & \ccwtab{$224^{2}$} & \ccwtab{80.0} & \ccwtab{CVPR'24}   \\
        & \ccwtab{GhostNetV3-1.3x~\cite{ghostnetv3}}$\dagger,\ddagger$             & \ccwtab{8.9}  & \ccwtab{\textcolor{gray}{269}}   & \ccwtab{$224^{2}$} & \ccwtab{79.1}  & \ccwtab{arXiv'2404}  \\
        & \ccwtab{MNetv4-Conv-M~\cite{mnetv4}}$\ast\dagger$& \ccwtab{9.2}  & \ccwtab{\textcolor{gray}{1000}}   & \ccwtab{$224^{2}$} & \ccwtab{79.9}  & \ccwtab{arXiv'2404}  \\
        & \ccgtab{DeiT-Ti~\cite{deit}}                     & \ccgtab{5.7}  & \ccgtab{1258} & \ccgtab{$224^{2}$} & \ccgtab{72.2} & \ccgtab{ICML'21}   \\
        & \ccgtab{XCiT-T12~\cite{xcit}}                    & \ccgtab{6.7}  & \ccgtab{1254} & \ccgtab{$224^{2}$} & \ccgtab{77.1} & \ccgtab{NeurIPS'21}\\
        & \ccgtab{LightViT-T~\cite{lightvit}}              & \ccgtab{9.4}  & \ccgtab{700}  & \ccgtab{$224^{2}$} & \ccgtab{78.7} & \ccgtab{arXiv'22}  \\
        & \ccgtab{MViTv1-S~\cite{mvitv1}}                  & \ccgtab{5.6}  & \ccgtab{2009} & \ccgtab{$256^{2}$} & \ccgtab{78.4} & \ccgtab{ICLR'22}   \\
        & \ccgtab{MViTv2-1.0~\cite{mvitv2}}                & \ccgtab{4.9}  & \ccgtab{1851} & \ccgtab{$256^{2}$} & \ccgtab{78.1} & \ccgtab{arXiv'22}  \\
        & \ccgtab{EdgeNeXt-S~\cite{edgenext}}              & \ccgtab{5.6}  & \ccgtab{965}  & \ccgtab{$224^{2}$} & \ccgtab{78.8} & \ccgtab{ECCVW'22}  \\
        & \ccgtab{PoolFormer-S12~\cite{metaformer}}        & \ccgtab{11.9} & \ccgtab{1823} & \ccgtab{$224^{2}$} & \ccgtab{77.2} & \ccgtab{CVPR'22}   \\
        & \ccgtab{MFormer-294M~\cite{mobileformer}}        & \ccgtab{11.4} & \ccgtab{294}  & \ccgtab{$224^{2}$} & \ccgtab{77.9} & \ccgtab{CVPR'22}   \\
        & \ccgtab{MPViT-T~\cite{mpvit}}                    & \ccgtab{5.8}  & \ccgtab{1654} & \ccgtab{$224^{2}$} & \ccgtab{78.2} & \ccgtab{CVPR'22}   \\
        & \ccgtab{EdgeViT-XS~\cite{edgevit}}               & \ccgtab{6.7}  & \ccgtab{1136} & \ccgtab{$256^{2}$} & \ccgtab{77.5} & \ccgtab{ECCV'22}   \\
        & \ccgtab{tiny-MOAT-1~\cite{moat}}                 & \ccgtab{5.1}  & \ccgtab{1200} & \ccgtab{$224^{2}$} & \ccgtab{78.3} & \ccgtab{ICLR'23}   \\
        & \ccgtab{EfficientViT-M5~\cite{evit}}             & \ccgtab{12.4} & \ccgtab{\textcolor{gray}{522}}  & \ccgtab{$224^{2}$} & \ccgtab{77.1} & \ccgtab{CVPR'23}   \\
        & \ccgtab{EfficientFormerV2-S1~\cite{efficientformerv2}}$\ast\dagger$             & \ccgtab{6.1}  & \ccgtab{\textcolor{gray}{650}}  & \ccgtab{$224^{2}$} & \ccgtab{79.0} & \ccgtab{ICCV'23}   \\
        & \ccgtab{ViG-T~\cite{vig}}             & \ccgtab{6.0}  & \ccgtab{\textcolor{gray}{900}}  & \ccgtab{$224^{2}$} & \ccgtab{77.2} & \ccgtab{arXiv'2405}   \\
        & \ccgtab{SHViT-S3~\cite{shvit}}             & \ccgtab{14.2}  & \ccgtab{\textcolor{gray}{601}}  & \ccgtab{$224^{2}$} & \ccgtab{77.4} & \ccgtab{CVPR'24}   \\
        & \ccgtab{EATFormer-Tiny~\cite{eatformer}}            & \ccgtab{6.1}  & \ccgtab{1410}  & \ccgtab{$224^{2}$} & \ccgtab{78.4}  & \ccgtab{IJCV'24} \\
        & \ccotab{Vim-Ti~\cite{vmamba}}             & \ccotab{7.0}  & \ccotab{\textcolor{gray}{1500}}  & \ccotab{$224^{2}$} & \ccotab{76.1} & \ccotab{ICML'24}   \\
        & \ccotab{EfficientVMamba-T~\cite{efficientvmamba}}             & \ccotab{6.0}  & \ccotab{\textcolor{gray}{800}}  & \ccotab{$224^{2}$} & \ccotab{76.5} & \ccotab{arXiv'2403}   \\
        & \ccotab{EfficientVMamba-S~\cite{efficientvmamba}}             & \ccotab{11.0}  & \ccotab{\textcolor{gray}{1300}}  & \ccotab{$224^{2}$} & \ccotab{78.7} & \ccotab{arXiv'2403}   \\
        & \ccotab{VRWKV-T~\cite{vrwkv}}             & \ccotab{6.2}  & \ccotab{\textcolor{gray}{1200}}  & \ccotab{$224^{2}$} & \ccotab{75.1} & \ccotab{arXiv'2403}   \\
        & \ccotab{MSVMamba-S~\cite{msvmamba}}             & \ccotab{7.0}  & \ccotab{\textcolor{gray}{900}}  & \ccotab{$224^{2}$} & \ccotab{77.3} & \ccotab{arXiv'2405}   \\
        & \ccotab{MambaOut-Femto~\cite{mambaout}}             & \ccotab{7.0}  & \ccotab{\textcolor{gray}{1200}}  & \ccotab{$224^{2}$} & \ccotab{78.9} & \ccotab{arXiv'2405}   \\
        & \ccbtab{\pzo\pzo~\ding{73}~EMOv1-5M~\cite{emo}}                         & \ccbtab{5.1}  & \ccbtab{903}  & \ccbtab{$224^{2}$} & \ccbtab{78.4} & \ccbtab{ICCV'23}   \\
        & \ccbtab{\ding{72}~\textbf{EMOv2-5M\pzo}}                        & \ccbtab{5.1}  & \ccbtab{1035} & \ccbtab{$224^{2}$} & \ccbtab{\textbf{79.4}} & \ccbtab{-} \\
        & \ccbtab{\ding{72}~\textbf{EMOv2-5M}}$\dagger$               & \ccbtab{5.1}  & \ccbtab{1035} & \ccbtab{$224^{2}$} & \ccbtab{\textbf{80.9}} & \ccbtab{-} \\
        & \ccbtab{\ding{72}~\textbf{EMOv2-5M}}$\ast$               & \ccbtab{5.1}  & \ccbtab{5627} & \ccbtab{$512^{2}$} & \ccbtab{\textbf{82.9}} & \ccbtab{-} \\
        \toprule[0.12em]
        \end{tabular}
    }
    \vspace{-1.0em}
\end{table}

\vspace{1mm} \noindent\textbf{Training recipes matters.} We evaluate EMO~\cite{emo} and EMOv2 with different mainstream training recipes presented in \cref{table:sota_recipe}. We find that our simple training recipe is enough to get impressive results, while existing stronger recipes (especially used by EdgeNeXt~\cite{edgenext}) will not improve performance further. \textit{NaN} indicates that the model did not train well for the possibly unadapted hyper-parameters. 

\subsection{Downstream Applications} \label{section:exp_downstream}
Thanks to the structural design of \textit{spanning attention} in \iirmb, our EMOv2 can simultaneously model global and local information interactions, which significantly enhances the performance of downstream tasks. It is noteworthy that current lightweight models have only reported limited results on downstream tasks, and different methods lack a unified experimental standard. Therefore, we have endeavored to find overlapping results from the original papers for a fair comparison. Additionally, we report the detailed results of our method with different magnitudes on multiple downstream tasks in the supplementary materials.

\begin{table}[tp]
    \centering
    \caption{Object detection performance by SSDLite~\cite{mnetv3} on MS-COCO 2017~\cite{coco} dataset at 320$\times$320 resolution. Abbreviated MNet/MViT: MobileNet/MobileViT. $\dagger$: 512 $\times$ 512 resolution.}
    \label{table:det_coco_ssdlite}
    \renewcommand{\arraystretch}{1.1}
    \setlength\tabcolsep{15.0pt}
    \resizebox{1.0\linewidth}{!}{
        \begin{tabular}{lccc}
        \toprule[0.17em]
        Backbone & \#Params $\downarrow$ & FLOPs $\downarrow$ & $mAP$ \\
        \hline
        MNetv1~\cite{mnetv1}        & 5.1       & 1.3G      & 22.2 \\
        MNetv2~\cite{mnetv2}        & 4.3       & 0.8G      & 22.1 \\
        MNetv3~\cite{mnetv3}        & 5.0       & 0.6G      & 22.0 \\
        MViTv1-XXS~\cite{mvitv1}    & 1.7       & 0.9G      & 19.9 \\
        MViTv2-0.5~\cite{mvitv2}    & 2.0       & 0.9G      & 21.2 \\
        \ding{73}~EMOv1-1M~\cite{emo}               & 2.3       & 0.6G      & 22.0 \\
        \ding{72}~\textbf{EMOv2-1M}             & 2.4       & 0.7G      & 22.3 \\
        \ding{72}~\textbf{EMOv2-1M}$\dagger$    & 2.4       & 2.3G      & 26.6 \\
        \hline
        MViTv2-0.75~\cite{mvitv2}   & 3.6       & 1.8G      & 24.6 \\
        \ding{73}~EMOv1-2M~\cite{emo}               & 3.3       & 0.9G      & 25.2\\
        \ding{72}~\textbf{EMOv2-2M}             & 3.3       & 1.2G      & 26.0\\
        \ding{72}~\textbf{EMOv2-2M}$\dagger$    & 3.3       & 4.0G      & 30.7\\
        \hline
        ResNet50~\cite{resnet}      & 26.6      & 8.8G      & 25.2 \\
        MViTv1-S~\cite{mvitv1}      & 5.7       & 3.4G      & 27.7 \\
        MViTv2-1.25~\cite{mvitv2}   & 8.2       & 4.7G      & 27.8 \\
        EdgeNeXt-S~\cite{edgenext}  & 6.2       & 2.1G      & 27.9 \\
        \ding{73}~EMOv1-5M~\cite{emo}               & 6.0       & 1.8G      & 27.9 \\
        \ding{72}~\textbf{EMOv2-5M}             & 6.0       & 2.4G      & 29.6 \\
        \ding{72}~\textbf{EMOv2-5M}$\dagger$    & 6.0       & 8.0G      & 34.8 \\
        \toprule[0.12em]
        \end{tabular}
    }
\end{table}

\begin{table}[tp]
    \centering
    \caption{Object detection results by RetinaNet~\cite{retinanet} on MS-COCO 2017~\cite{coco} dataset.}
    \label{table:det_coco_retinanet}
    \renewcommand{\arraystretch}{1.2}
    \setlength\tabcolsep{3.0pt}
    \resizebox{1.\linewidth}{!}{
        \begin{tabular}{lccccccc}
        \toprule[0.17em]
        Backbone & \#Params & $mAP^b$  & $mAP^b_{50}$ & $mAP^b_{75}$ & $mAP^b_{S}$ & $mAP^b_{M}$ & $mAP^b_{L}$ \\
        \hline
        ResNet-50~\cite{resnet}     & 37.7 & 36.3 & 55.3 & 38.6 & 19.3 & 40.0 & 48.8 \\
        PVTv1-Tiny~\cite{pvtv1}     & 23.0 & 36.7 & 56.9 & 38.9 & 22.6 & 38.8 & 50.0 \\
        PVTv2-B0~\cite{pvtv2}       & 13.0 & 37.2 & 57.2 & 39.5 & 23.1 & 40.4 & 49.7 \\
        EdgeViT-XXS~\cite{edgevit}  & 13.1 & 38.7 & 59.0 & 41.0 & 22.4 & 42.0 & 51.6 \\
        \ding{73}~EMOv1-5M            & 14.4 & 38.9 & 59.8 & 41.0 & 23.8 & 42.2 & 51.7 \\
        \ding{72}~\textbf{EMOv2-5M} & 14.4 & 41.5 & 62.7 & 44.1 & 25.7 & 45.5 & 55.5 \\
        \toprule[0.12em]
        \end{tabular}
    }
    \vspace{-1.0em}
\end{table}

\vspace{1mm} \noindent\textbf{Object detection.} 
We evaluate our EMOv2 (pre-trained on ImageNet-1K) with other SoTA methods on MS-COCO 2017~\cite{coco} dataset, using the lightweight SSDLite~\cite{mnetv3} and heavy RetinaNet~\cite{retinanet} / Mask RCNN~\cite{maskrcnn}. Considering fairness and friendliness for the community, we employ standard MMDetection library~\cite{mmdetection} for experiments and replace the optimizer with AdamW~\cite{adamw} without tuning other parameters. 

Comparison results on SSDLite are shown in \cref{table:det_coco_ssdlite}, and our EMOv1surpasses corresponding counterparts by apparent advantages and the improved EMOv2 further boosts the performance. 
For example, SSDLite equipped with EMOv1-1M achieves 22.0 mAP with only 0.6G FLOPs and 2.3M parameters, which boosts +2.1$\uparrow$ compared with SoTA MobileViT~\cite{mvitv1} with only 66\% FLOPs. Consistently, EMOv1-5M obtains the highest 27.9 mAP so far with much fewer FLOPs, \eg, 53\% (1.8G) of MobileViT-S~\cite{mvitv1} (3.4G) and 0.3G less than EdgeNeXt-S (2.1G). EMOv2-5M further achieves 29.6 mAP with no significant increase in parameters, surpassing EMOv1-5M by +1.7$\uparrow$. 
We also conduct experiments on heavy detection frameworks. \cref{table:det_coco_retinanet} and \cref{table:det_coco_maskrcnn} present the results of different lightweight backbones on the RetinaNet~\cite{retinanet} and Mask RCNN~\cite{maskrcnn} methods, respectively. Our EMOv2 consistently achieves superior results compared to its counterparts, \eg, +5.2$\uparrow$ $mAP$ over the CNN-based ResNet-50, +2.8$\uparrow$ $mAP$ over the Transformer-based EdgeViT-XXS, and +2.6$\uparrow$ $mAP$ over our previous EMOv1under the RetinaNet framework. For the Mask RCNN framework, our EMOv2-5M obtains highly competitive results compared to the recently designed EATFormer for heavy architectures, with improvements of +3.0$\uparrow$ $mAP^b$ and +2.6$\uparrow$ $mAP^m$ over the previous generation EMOv1-5M model.

\begin{table}[tp]
    \centering
    \caption{Object detection results by Mask RCNN~\cite{maskrcnn} on MS-COCO 2017~\cite{coco} dataset.}
    \label{table:det_coco_maskrcnn}
    \renewcommand{\arraystretch}{1.2}
    \setlength\tabcolsep{3.0pt}
    \resizebox{1.\linewidth}{!}{
        \begin{tabular}{lccccccc}
        \toprule[0.17em]
        \multirow{2}{*}{Backbone} & \multirow{2}{*}{\#Params $\downarrow$} & $mAP^b$  & $mAP^b_{50}$ & $mAP^b_{75}$ & $mAP^b_{S}$ & $mAP^b_{M}$ & $mAP^b_{L}$ \\
        \cline{3-8}
        &  &  $mAP^m$  & $mAP^m_{50}$ & $mAP^m_{75}$ & $mAP^m_{S}$ & $mAP^m_{M}$ & $mAP^m_{L}$ \\
        \hline
        \multirow{2}{*}{PVT-Tiny~\cite{pvtv1}} & \multirow{2}{*}{33.0} & 36.7 & 59.2 & 39.3 & - & - & - \\
        \cline{3-8}
        & & 35.1 & 56.7 & 37.3 & - & - & - \\

        \hline
        \multirow{2}{*}{PVTv2-B0~\cite{pvtv2}} & \multirow{2}{*}{23.0} & 38.2 & 60.5 & 40.7 & - & - & - \\
        \cline{3-8}
        & & 36.2 & 57.8 & 38.6 & - & - & - \\

        \hline
        \multirow{2}{*}{PoolFormer-S12~\cite{metaformer}} & \multirow{2}{*}{31.0} & 37.3 & 59.0 & 40.1 & - & - & - \\
        \cline{3-8}
        & & 34.6 & 55.8 & 36.9 & - & - & - \\

        \hline
        \multirow{2}{*}{MPViT-T~\cite{mpvit}} & \multirow{2}{*}{28.0} & 42.2 & 64.2 & 45.8 & - & - & - \\
        \cline{3-8}
        & & 39.0 & 61.4 & 41.8 & - & - & - \\

        \hline
        \multirow{2}{*}{EATFormer-Tiny~\cite{eatformer}} & \multirow{2}{*}{25.9} & 42.3 & 64.7 & 46.2 & 25.5 & 45.5 & 55.1 \\
        \cline{3-8}
        & & 39.0 & 61.5 & 42.0 & 22.4 & 42.0 & 52.7 \\
        
        \hline
        \multirow{2}{*}{\ding{73}~EMOv1-5M} & \multirow{2}{*}{24.8} & 39.3 & 61.7 & 42.4 & 23.5 & 42.3 & 51.1 \\
        \cline{3-8}
        & & 36.4 & 58.4 & 38.7 & 18.2 & 39.0 & 52.6 \\

        \hline
        \multirow{2}{*}{\ding{72}~\textbf{EMOv2-5M}} & \multirow{2}{*}{24.8} & 42.3 & 64.3 & 46.3 & 25.8 & 45.6 & 56.3 \\
        \cline{3-8}
        & & 39.0 & 61.4 & 42.1 & 20.0 & 41.8 & 57.0 \\
        \toprule[0.12em]
        \end{tabular}
    }
\end{table}

\vspace{1mm} \noindent\textbf{Semantic segmentation.} 
ImageNet-1K pre-trained EMOv2 is integrated with DeepLabv3~\cite{deeplabv3}, Semantic FPN~\cite{semanticfpn}, SegFormer~\cite{segformer}, and PSPNet~\cite{pspnet} to adequately evaluate its performance on challenging ADE20K~\cite{ade20k} dataset at 512$\times$512 resolution. We employ the standard MMSegmentation library~\cite{mmseg2020} with official configurations without tuning other parameters. 

Due to the fact that different methods only report results on certain segmentation frameworks, we strive to find sufficient comparable models of similar magnitude under each method. Detailed results are presented in \cref{table:seg_four}. For lightweight models at the 1M/2M/5M magnitude, our method demonstrates significant advantages over comparative methods (including CNN, Transformer, and hybrid architectures), achieving a balance between parameters, computational cost, and performance. Notably, our conference version model (\ie, EMO~\cite{emo}) achieves highly competitive results, and the improved EMOv2 model further significantly enhances the metrics. For instance, under the Deeplabv3 framework, our EMOv2-1M/2M/5M achieved 34.6/36.8/39.8 mIoU, respectively, representing improvements of +1.1$\uparrow$/+1.5$\uparrow$/+2.0$\uparrow$ over EMOv1with fewer parameters. Similarly, under the Semantic FPN framework, our EMOv2-1M/2M/5M achieves 37.1/39.9/42.3 mIoU, respectively, representing improvements of +2.9$\uparrow$/+2.6$\uparrow$/+1.9$\uparrow$ over EMOv1without increasing the number of parameters. More detailed results can be found in the supplementary materials.

\begin{table}[tp]
    \centering
    \caption{Semantic segmentation results by DeepLabv3~\cite{deeplabv3}, Semantic FPN~\cite{semanticfpn}, SegFormer~\cite{segformer}, and PSPNet~\cite{pspnet} on ADE20K~\cite{ade20k} dataset at 512$\times$512 resolution.}
    \label{table:seg_four}
    \renewcommand{\arraystretch}{1.0}
    \setlength\tabcolsep{10.0pt}
    \resizebox{1.0\linewidth}{!}{
        \begin{tabular}{clccc}
        \toprule[0.17em]
        & Backbone & \#Params $\downarrow$ & FLOPs $\downarrow$ & mIoU \\
        \hline
        \multirow{13}{*}{\rotatebox{90}{\makecell[c]{DeepLabv3\\~\cite{deeplabv3}}}} 
        & MViTv2-0.5 & 6.3 & 26.1G & 31.9 \\ 
        & MViTv3-0.5 & 6.3 & - & 33.5 \\ 
        & \ding{73}~EMOv1-1M & 5.6 & 2.4G & 33.5 \\ 
        & \ding{72}~\textbf{EMOv2-1M} & 5.6 & 3.3G & 34.6 \\ 
        \cline{2-5}
        & MNetv2 & 18.7 & 75.4G & 34.1 \\ 
        & MViTv2-0.75 & 9.6 & 40.0G & 34.7 \\ 
        & MViTv3-0.75 & 9.7 & - & 36.4 \\ 
        & \ding{73}~EMOv1-2M & 6.9 & 3.5G & 35.3 \\ 
        & \ding{72}~\textbf{EMOv2-2M} & 6.6 & 5.0G & 36.8 \\ 
        \cline{2-5}
        & MViTv2-1.0 & 13.4 & 56.4G & 37.0 \\ 
        & MViTv3-1.0 & 13.6 & - & 39.1 \\ 
        & \ding{73}~EMOv1-5M & 10.3 & 5.8G & 37.8 \\ 
        & \ding{72}~\textbf{EMOv2-5M} & 9.9 & 9.1G & 39.8 \\ 
        \hline
        \multirow{16}{*}{\rotatebox{90}{\makecell[c]{Semantic FPN\\~\cite{semanticfpn}}}} 
        & ResNet-18 & 15.5 & 32.2G & 32.9 \\ 
        & \ding{73}~EMOv1-1M & 5.2 & 22.5G & 34.2 \\ 
        & \ding{72}~\textbf{EMOv2-1M} & 5.3 & 23.4G & 37.1 \\ 
        \cline{2-5}
        & ResNet-50 & 28.5 & 45.6G & 36.7 \\ 
        & PVTv1-Tiny & 17.0 & 33.2G & 35.7 \\ 
        & PVTv2-B0 & 7.6 & 25.0G & 37.2 \\ 
        & \ding{73}~EMOv1-2M & 6.2 & 23.5G & 37.3 \\ 
        & \ding{72}~\textbf{EMOv2-2M} & 6.2 & 25.1G & 39.9 \\ 
        \cline{2-5}
        & ResNet-101 & 47.5 & 65.1G & 38.8 \\ 
        & ResNeXt-101 & 47.1 & 64.7G & 39.7 \\ 
        & PVTv1-Small & 28.2 & 44.5G & 39.8 \\ 
        & EdgeViT-XXS & 7.9 & 24.4G & 39.7 \\ 
        & EdgeViT-XS & 10.6 & 27.7G & 41.4 \\ 
        & PVTv2-B1 & 17.8 & 34.2G & 42.5 \\ 
        & \ding{73}~EMOv1-5M & 8.9 & 25.8G & 40.4 \\ 
        & \ding{72}~\textbf{EMOv2-5M} & 8.9 & 29.1G & 42.3 \\ 
        \hline
        \multirow{4}{*}{\rotatebox{90}{\makecell[c]{SegFormer\\~\cite{segformer}}}} 
        & MiT-B0 & 3.8 & 8.4G & 37.4 \\ 
        & \ding{72}~\textbf{EMOv2-2M} & 2.6 & 10.3G & 40.2 \\ 
        \cline{2-5}
        & MiT-B1 & 13.7 & 15.9G & 42.2 \\ 
        & \ding{72}~\textbf{EMOv2-5M} & 5.3 & 14.4G & 43.0 \\ 
        \hline
        \multirow{10}{*}{\rotatebox{90}{\makecell[c]{PSPNet\\~\cite{pspnet}}}} 
        & MNetv2 & 13.7 & 53.1G & 29.7 \\ 
        & MViTv2-0.5 & 3.6 & 15.4G & 31.8 \\ 
        & \ding{73}~EMOv1-1M & 4.3 & 2.1G & 33.2 \\ 
        & \ding{72}~\textbf{EMOv2-1M} & 4.2 & 2.9G & 33.6 \\ 
        \cline{2-5}
        & MViTv2-0.75 & 6.2 & 26.6G & 35.2 \\ 
        & \ding{73}~EMOv1-2M & 5.5 & 3.1G & 34.5 \\ 
        & \ding{72}~\textbf{EMOv2-2M} & 5.2 & 4.6G & 35.7 \\ 
        \cline{2-5}
        & MViTv2-1.0 & 9.4 & 40.3G & 36.5 \\ 
        & \ding{73}~EMOv1-5M & 8.5 & 5.3G & 38.2 \\ 
        & \ding{72}~\textbf{EMOv2-5M} & 8.1 & 8.6G & 39.1 \\ 
        \bottomrule[0.12em]
        \end{tabular}
    }
\end{table}

\begin{table}[tp]
    \centering
    \caption{Semantic segmentation results by UNet~\cite{unet} on HRF~\cite{hrf} dataset at 256$\times$256 resolution.}
    \label{table:seg_unet}
    \renewcommand{\arraystretch}{1.1}
    \setlength\tabcolsep{10.0pt}
    \resizebox{1.0\linewidth}{!}{
        \begin{tabular}{lccccc}
        \toprule[0.17em]
        Backbone & \#Params $\downarrow$ & FLOPs $\downarrow$ & mDice & aAcc & mAcc \\
        \hline
        UNet-S5-D16 & 29.0 & 204G & 88.9 & 97.0 & 86.2 \\ 
        EdgeNeXt-S~\cite{edgenext} & 23.7 & 221G & 89.1 & 97.1 & 87.5 \\ 
        \ding{72}~\textbf{U-EMOv2-5M} & 21.3 & 228G & 89.5 & 97.1 & 88.3 \\ 
        \bottomrule[0.12em]
        \end{tabular}
    }
\end{table}

Previous studies have demonstrated the effectiveness of EMOv2 in classification and mainstream downstream detection/segmentation tasks. To further validate the superiority of EMOv2, we additionally extend it to UNet-like architectures, as well as video classification and DiT-based image generation. 

\vspace{1mm} \noindent\textbf{UNet-based vision segmentation (U-EMO).} 
Furthermore, we replace the basic convolutional block in UNet with the {\iirmb} block to construct a more powerful U-EMO architecture, as described in \cref{fig:emo2}, and we conduct experiments on the downstream segmentation task to demonstrate the generalizability of the proposed method across different architectures. \cref{table:seg_unet} presents results of U-EMO, UNet~\cite{unet}, and the adapted EdgeNeXt~\cite{edgenext} method on the HRF~\cite{hrf} dataset at 256$\times$256 resolution. Our improved U-EMO achieves higher performance with fewer parameters without meticulous adjustments to the architecture and training recipes. 

\vspace{1mm} \noindent\textbf{Video classification (V-EMO).} 
By simply extending the temporal dimension of the convolution and spanning attention in the {\iirmb} block, we obtain a basic \iirmb-3D block for video processing. This allows us to replace modules while maintaining a structure similar to 2D EMOv2, resulting in the V-EMO model. We use ImageNet-1K pretrained weights with temporal repetition to initialize the video classification model. \cref{table:video} presents a comparison of our method with UniFormer-XXS~\cite{uniformer} and the adapted EdgeNeXt~\cite{edgenext} method on the Kinetics-400~\cite{k400} dataset. Our V-EMO-5M achieves a Top-1 accuracy of 65.2 with only 5.9M parameters, outperforming UniFormer-XXS, which has 9.8M parameters, by +2.0$\uparrow$. 

\begin{table}[tp]
    \centering
    \caption{Comparison with the state-of-the-art on Kinetics-400~\cite{k400} dataset with four input frames.}
    \label{table:video}
    \renewcommand{\arraystretch}{1.2}
    \setlength\tabcolsep{15.0pt}
    \resizebox{1.0\linewidth}{!}{
        \begin{tabular}{lccc}
        \toprule[0.17em]
        Backbone & \#Params $\downarrow$ & FLOPs $\downarrow$ & Top-1 \\
        \hline
        UniFormer-XXS & 9.8 & 1.0G & 63.2 \\ 
        EdgeNeXt-S~\cite{edgenext} & 6.8 & 1.2G & 64.3 \\ 
        \ding{72}~\textbf{V-EMOv2-5M} & 5.9 & 1.3G & 65.2 \\ 
        \bottomrule[0.12em]
        \end{tabular}
    }
\end{table}

\begin{table}[tp!]
    \centering
    \caption{Comparison with DiT~\cite{dit} for 400K training steps in generating 256$\times$256 ImageNet~\cite{imagenet} images. }
    \label{table:dit}
    \renewcommand{\arraystretch}{1.1}
    \setlength\tabcolsep{15.0pt}
    \resizebox{1.0\linewidth}{!}{
        \begin{tabular}{rccc}
        \toprule[0.17em]
        Model & \#Params $\downarrow$ & FLOPs $\downarrow$ & FID \\
        \hline
        DiT-S-2 & \pzo33.0 & \pzo\pzo5.5G & 68.4 \\ 
        SiT-S-2 & \pzo33.0 & \pzo\pzo5.5G & 57.6 \\ 
        D-EMOv2-S-2 & \pzo24.6 & \pzo\pzo5.4G & 46.3 \\ 
        \hline
        DiT-B-2 & 130.5 & \pzo21.8G & 43.5 \\ 
        SiT-B-2 & 130.5 & \pzo21.8G & 33.5 \\ 
        D-EMOv2-B-2 & \pzo96.1 & \pzo19.9G & 24.8 \\ 
        \hline
        DiT-L-2 & 458.1 & \pzo77.5G & 23.3 \\ 
        SiT-L-2 & 458.1 & \pzo77.5G & 18.8 \\ 
        D-EMOv2-L-2 & 334.8 & \pzo69.3G & 11.2 \\ 
        \hline
        DiT-XL-2 & 675.1 & 114.5G & 19.5 \\ 
        SiT-XL-2 & 675.1 & 114.5G & 17.2 \\ 
        D-EMOv2-XL-2 & 492.7 & 101.5G & \pzo9.6 \\ 
        \bottomrule[0.12em]
        \end{tabular}
    }
\end{table}

\vspace{1mm} \noindent\textbf{DiT-based image generation (D-EMO).} 
The primary design goal of the {\iirmb} is to simplify the Transformer block structure, making it suitable for mobile architecture design by reducing the depth of individual blocks while improving the modeling of both distant and neighboring features. Thanks to its plug-and-play characteristic, {\iirmb} can easily replace the Transformer block in the DiT model for image generation tasks. Specifically, we fully adhere to the DiT~\cite{dit} training framework, and the results on the 256$\times$256 ImageNet generation task are shown in \cref{table:dit}. Compared to the baseline DiT~\cite{dit} and the SiT~\cite{sit} with improved training strategies, our D-EMO model, which replaces the basic Transformer block with \iirmb, requires fewer parameters and computational resources while achieving significantly better FID scores. This demonstrates the advantage of spanning attention in downstream image generation task.

\subsection{Structural Ablation and Analysis} \label{section:exp_ablation}
This section uses EMOv2-5M as the research backbone to ablate the proposed method modules and training hyperparameters, while also analyzing the model structure and results. 

\vspace{1mm} \noindent\textbf{Depth and channel configurations.} 
Using EMOv2-5M as the baseline, we evaluate the impact of different depth configurations on model performance, as shown in the upper part of \cref{table:exp_depth}. The selected depth configuration yields a relatively better performance. Furthermore, we assess the performance of slimmer and wider models with a similar number of parameters, as shown in the lower part of \cref{table:exp_depth}. These models, despite having an increased computational load, do not result in further performance improvements, demonstrating the rationality of the current structural configuration.

\begin{table}[tp]
    \centering
    \caption{Efficiency and performance comparison of different depth and channel configurations.
    }
    \label{table:exp_depth}
    \renewcommand{\arraystretch}{1.1}
    \setlength\tabcolsep{5.0pt}
    \resizebox{0.9\linewidth}{!}{
        \begin{tabular}{ccccc}
        \toprule[0.2em]
        $~$Depth & Channels & \#Params & FLOPs & Top-1 \\
        \hline
        $~$[2, 2, 10, 3] & [48, 72, 160, 288]  & 5.3M & 1038M & 79.1 \\
        $~$[2, 2, 12, 2] & [48, 72, 160, 288] & 5.0M & 1127M & 78.9 \\
        $~$[4, 4, 8, 3] & [48, 72, 160, 288] & 5.1M & 1132M & 79.4 \\
        \hline
        $~$[3, 3, 9, 3] & [48, 72, 160, 288] & 5.1M & 1035M & 79.4 \\
        \hline
        $~$[2, 2, 12, 3] & [48, 72, 160, 288] & 5.1M & 1136M & 79.1 \\
        $~$[2, 2, 8, 2] & [48, 72, 224, 288]  & 5.1M & 1117M & 79.0 \\
        \toprule[0.2em]
        \end{tabular}
    }
\end{table}

\vspace{1mm} \noindent\textbf{Throughput comparison.} 
\cref{table:exp_throughput} presents throughput evaluation results compared with the state-of-the-art EdgeNeXt~\cite{edgenext}, which effectively balances parameters, computational load, and performance. The test platforms are an AMD EPYC 7K62 CPU and a V100 GPU, with a resolution of 224$\times$224 and a batch size of 256. Results indicate that EMOv1achieves faster speeds on both platforms with higher Top-1 accuracy. For instance, EMOv1-1M achieves speed boosts of +20\%$\uparrow$ on the GPU and +116\%$\uparrow$ on the CPU compared to EdgeNeXt-XXS with the same FLOPs. 
The improved EMOv2 maintains nearly the same parameter count as EMOv1but significantly enhances performance with a slight increase in computational load. This performance gap is further widened on mobile devices (following the official classification project~\cite{ios_cls} on iPhone15), where our EMOv2 is 2.8$\times\uparrow$, 4.1$\times\uparrow$, and 3.9$\times\uparrow$ faster than the state-of-the-art EdgeNeXt~\cite{edgenext}. This improvement is attributed to our simple and device-friendly {\iirmb} block, which does not rely on other complex structures such as the Res2Net module~\cite{res2net}, transposed channel attention~\cite{xcit}, \etc

\begin{table}[tp]
    \centering
    \caption{Comparisons of throughput on CPU/GPU and running speed on mobile iPhone15 (ms).}
    \label{table:exp_throughput}
    \renewcommand{\arraystretch}{1.2}
    \setlength\tabcolsep{6.0pt}
    \resizebox{1.0\linewidth}{!}{
        \begin{tabular}{lcccccc}
        \toprule[0.2em]
        \pzo\pzo Method & \#Params $\downarrow$ & FLOPs & CPU & GPU & iPhone15 & Top-1 \\
        \hline
        EdgeNeXt-XXS    & 1.3M & 261M & \pzo73.1   & 2860.6    & 10.2 & 71.2 \\
        \ding{73}~\textbf{EMOv1-1M} & 1.3M & 261M & 158.4  & 3414.6    & \pzo3.0 & 71.5 \\
        \ding{72}~\textbf{EMOv2-1M} & 1.4M & 285M & 147.1  & 3182.2    & \pzo3.6 & 72.3 \\
        \hline
        EdgeNeXt-XS     & 2.3M & 538M & \pzo69.1   & 1855.2    & 17.6 & 75.0 \\
        \ding{73}~\textbf{EMOv1-2M} & 2.3M & 439M & 126.6  & 2509.8    & \pzo3.7 & 75.1 \\
        \ding{72}~\textbf{EMOv2-2M} & 2.3M & 487M & 118.2  & 3312.4    & \pzo4.3 & 75.8 \\
        \hline
        EdgeNeXt-S      & 5.6M & 965M & \pzo54.2   & 1622.5    & 22.5 & 78.8 \\
        \ding{73}~\textbf{EMOv1-5M} & 5.1M & 903M & 106.5  & 1731.7    & \pzo4.9 & 78.4 \\
        \ding{72}~\textbf{EMOv2-5M} & 5.1M & 1035M & 93.9  & 1607.8    & \pzo5.9 & 79.4 \\
        \toprule[0.2em]
        \end{tabular}
    }
\end{table}

\begin{table*}[tp]
    \centering
    \caption{Ablation studies and comparison analysis on ImageNet~\cite{imagenet}. All the experiments use EMOv2-5M as default structure.}
    \label{table:ablation_all}
    \subfloat[Attention mode analysis on classification and downstream RetinaNet~\cite{retinanet} / DeepLabv3~\cite{deeplabv3}.]{
        \label{tab:ablation_a}
        \renewcommand{\arraystretch}{1.25}
        \setlength\tabcolsep{3.0pt}
        \resizebox{0.42\linewidth}{!}{
        \begin{tabular}{cccccc}
        \toprule[0.15em]
        Mode & \#Params $\downarrow$ & FLOPs $\downarrow$ & Top-1 & mAP & mIoU \\
        \hline
        None & 4.3M & 802M & 77.9 & 39.3 & 37.2 \\ 
        None (Scaling to 5.1M) & 5.1M & 991M & 78.4 & 39.6 & 37.7 \\ 
        Neighborhood Attention & 5.1M & 967M & 78.8 & 40.4 & 39.0 \\ 
        Remote Attention & 5.1M & 967M & 79.0 & 39.9 & 38.6 \\ 
        Spanning Attention & 5.1M & 1035M & 79.4 & 41.5 & 39.8 \\ 
        \bottomrule[0.12em]
        \end{tabular}
    }
    } \hfill
    \subfloat[Applied stages of spanning attention.]{
        \label{tab:ablation_b}
        \renewcommand{\arraystretch}{1.2}
        \setlength\tabcolsep{3.0pt}
        \resizebox{0.29\linewidth}{!}{
        \begin{tabular}{lccc}
        \toprule[0.10em]
        Stage & \#Params $\downarrow$ & FLOPs $\downarrow$ & Top-1 \\
        \hline
        S-4 & 4.7M & 832M & 78.5 \\
        S-34 & 5.1M & 1035M & 79.4 \\ 
        S-234 & 5.1M & 1096M & 79.3 \\ 
        S-1234 & 5.2M & 1213M & 79.1 \\ 
        \bottomrule[0.10em]
        \end{tabular}
    }
    } \hfill
    \subfloat[Influence of DPR and BS hyperparameters.]{
        \label{tab:ablation_c}
        \renewcommand{\arraystretch}{0.96}
        \setlength\tabcolsep{3.0pt}
        \resizebox{0.22\linewidth}{!}{
        \begin{tabular}{cc|cc}
        \toprule[0.12em]
        DPR & Top-1 & BS & Top-1 \\
        \hline
        0.00 & 79.1 & 256 & 78.9 \\
        0.03 & 79.2 & 512 & 79.2 \\
        0.05 & 79.4 & 1024 & 79.4 \\
        0.10 & 79.3 & 2048 & 79.4 \\
        0.20 & 79.1 & 4096 & 79.4 \\
        \bottomrule[0.12em]
        \end{tabular}
    }
    } \hfill
    \vspace{0.5em}
    \subfloat[Convolution type. K: kernel size. D: Dilation.]{
        \label{tab:ablation_d}
        \renewcommand{\arraystretch}{1.3}
        \setlength\tabcolsep{9.0pt}
        \resizebox{0.33\linewidth}{!}{
        \begin{tabular}{cccc}
        \toprule[0.15em]
        Size & \#Params $\downarrow$ & FLOPs $\downarrow$ & Top-1 \\
        \hline
        K-1 & 4.8M & 969M & 78.6 \\
        K-3 & 4.9M & 991M & 79.0 \\
        K-5 & 5.1M & 1035M & 79.4 \\
        K-7 & 5.3M & 1102M & 79.2 \\
        K-9 & 5.5M & 1184M & 79.3 \\
        \hline
        K-5 + D-2 & 5.1M & 1035M & 79.3 \\
        K-5 + D-3 & 5.1M & 1035M & 79.1 \\
        K-5 + DCNv2~\cite{dcnv2}  & 6.7M & 1625M & 78.5 \\
        \bottomrule[0.15em]
        \end{tabular}
    }
    } \hfill
    \subfloat[Training strategies: image resolution, knowledge distillation, and 1000 training epochs.]{
        \label{tab:ablation_e}
        \renewcommand{\arraystretch}{1.0}
        \setlength\tabcolsep{10.0pt}
        \resizebox{0.6\linewidth}{!}{
        \begin{tabular}{cccccc}
        \toprule[0.10em]
        Resolution & KD & Long Training & \#Params. & FLOPs & Top-1 \\
        \hline
        224 & \xmark & \xmark & 1.0G & 5.1M & 79.4 \\
        \hline
        256 & \xmark & \xmark & 1.4G & 5.1M & 79.9 \\
        224 & \cmark & \xmark & 1.0G & 5.1M & 80.8 \\
        224 & \xmark & \cmark & 1.0G & 5.1M & 80.4 \\
        \hline
        512 & \xmark & \xmark & 5.6G & 5.1M & 81.5 \\
        512 & \cmark & \xmark & 5.6G & 5.1M & 82.4 \\
        512 & \cmark & \cmark & 5.6G & 5.1M & 82.9 \\
        \bottomrule[0.12em]
        \end{tabular}
    }
    } \hfill
\end{table*}

\vspace{1mm} \noindent\textbf{Attention mode.} 
The proposed {\iirmb} in \cref{section:iirmb} includes two components: distant and neighbor window attention with shared parameters. \cref{tab:ablation_a} evaluates the model's performance under different attention modes. When neighborhood and distant attention are added separately, the model shows significant improvement compared to the baseline model. It also outperforms models of similar magnitude without attention, especially in downstream task metrics, demonstrating the effectiveness of the proposed basic EW-MHSA (\cref{section:irmb}). Thanks to the shared parameter design, the model with integrated spanning attention achieves better Top-1 classification results without any additional parameters. This is particularly evident in detection and segmentation tasks, further proving the effectiveness of the spanning mechanism in \iirmb. 

\vspace{1mm} \noindent\textbf{Used stages of spanning attention.} 
\cref{tab:ablation_b} shows the changes in model accuracy when applying spanning attention to different stages based on EMOv2-5M. As spanning attention is gradually added from the fourth stage (S-4) to all four stages (S-1234), the model's performance significantly increases (S-34) and then saturates and slightly decreases (S-234). Considering that more stages require additional parameters and computational resources, spanning attention is by default injected only in the last two stages. Interestingly, in the conference version of EMO~\cite{emo}, the accuracy of the model increases with the number of stages to which spanning attention is applied. This discrepancy may be due to the structure of \iirmb, where EMOv2-5M is closer to the performance upper limit for models with this parameter count.

\vspace{1mm} \noindent\textbf{Effect of training hyper-parameters.} 
\cref{tab:ablation_c} discusses the two most influential hyperparameters in model training. The proposed EMOv2-5M exhibits strong robustness to the drop path rate (DPR) hyperparameter within the range of [0, 0.2], where the Top-1 accuracy fluctuates within 0.3, achieving the best result at a drop path rate of 0.05. Meanwhile, a smaller batch size (BS) of 256 slightly affects the model's performance, with the performance peaking at a batch size of 1024 and then stabilizing. Considering memory efficiency, a default batch size of 1024 is suggested. These ablation experiments demonstrate the robustness of EMOv2 to the above hyperparameter variations.

\vspace{1mm} \noindent\textbf{Neighborhood kernel size in \iirmb.} 
The size of the DW-Conv affects the local receptive field of \iirmb, which significantly impacts the model's classification ability and perception capability in downstream tasks. As shown in \cref{tab:ablation_d}-Top, when the kernel size gradually increases from 1 to 5, the model's performance improves from 78.6 to 79.4. However, further increases in kernel size do not yield noticeable gains and instead incur additional parameter and computational costs.

\vspace{1mm} \noindent\textbf{Convolution type in \iirmb.} 
\cref{tab:ablation_d}-Bottom illustrates the impact of different convolution variants on EMOv2, which extend the receptive field. The use of dilated convolutions does not further improve the model's performance; in fact, when the dilation rate is set to 3, the model's performance slightly decreases. Deformable convolution significantly increases the model's parameter count and computational load. Therefore, we replace the DW-Conv in EMOv2-1M with DCNv2~\cite{dcnv2} with a group size of 1 to maintain a similar scale of the model. The results indicate that this substitution actually reduces the model's performance.

\vspace{1mm} \noindent\textbf{Stronger training strategy.}
\cref{tab:ablation_e} presents three training strategies that enhance model performance without altering the model architecture or parameters. When employing higher resolutions (up to 512 in this paper), knowledge distillation (KD) with naive logit distribution (TResNet~\cite{tresnet} in \cref{section:exp_cls}), and long training durations (up to 1000 epochs), the model's performance improves significantly. When all strategies are combined, the EMOv2-5M achieves the best 82.9 Top-1 accuracy. This performance notably surpasses that of Swin-Transformer-T (28.2M with 81.3 Top-1) and ResNet-152 (60.1M with 82.0 Top-1).

\vspace{1mm} \noindent\textbf{Scale up assessment}
We scale up EMOv2 to 20M/50M magnitudes to evaluate its scaling capability. The specific structure is presented in \cref{table:model_variants_scale_up}, and the comparison results with current backbones of similar magnitudes are shown in \cref{table:cls_imagenet_supp}. The results demonstrate that EMOv2 can be easily extended to large-scale models and achieve highly competitive results. This scaling capability is also reflected in \cref{table:dit}, proving the structural effectiveness and generalization of \iirmb. 

\begin{table}[tp]
    \centering
    \caption{Core configurations of scaled EMOv2 variants.}
    \label{table:model_variants_scale_up}
    \renewcommand{\arraystretch}{1.2}
    \setlength\tabcolsep{10.0pt}
    \resizebox{0.9\linewidth}{!}{
        \begin{tabular}{ccc}
            \toprule[0.17em]
            Items           & EMOv2-20M                 & EMOv2-50M \\
            \hline
            Depth           & [ 3, 3, 13, 3 ]           & [ 5, 8, 20, 7 ]  \\
            Emb. Dim.       & [ 64, 128, 320, 448 ]     & [ 64, 128, 384, 512 ] \\
            Exp. Ratio      & [ 2.0, 3.0, 4.0, 4.0 ]        & [ 2.0, 3.0, 4.0, 4.0 ] \\
            \toprule[0.12em]
        \end{tabular}
    }
\end{table}

\begin{table}[tp]
    \centering
    \caption{Evaluation of scaling capabilities of EMOv2 at 20M/50M magnitudes on ImageNet-1K dataset.
    } 
    \label{table:cls_imagenet_supp}
    \renewcommand{\arraystretch}{1.1}
    \setlength\tabcolsep{5.0pt}
    \resizebox{1.\linewidth}{!}{
        \begin{tabular}{clccccc}
        \toprule[0.17em]
        & Model & \#Params $\downarrow$ & FLOPs $\downarrow$ & Reso. & Top-1 & Venue \\
        \hline
        \multirow{9}{*}{\rotatebox{90}{\makecell[c]{\textbf{20M} \\ \textbf{Magnitude}}}}
        & ResNet-50~\cite{resnet,timm_resnet} & 25.5      & 4.1G      & $224^{2}$ & 80.4  & CVPR'16   \\
        & ConvNeXt-T~\cite{convnet}           & 28.5      & 4.5G      & $224^{2}$ & 82.1  & CVPR'22   \\
        & PVTv2-B2~\cite{pvtv2}               & 25.3      & 4.0G      & $224^{2}$ & 82.0  & ICCV'21   \\
        & Swin-T~\cite{swinv1}                & 28.2      & 4.5G      & $224^{2}$ & 81.3  & ICCV'21   \\
        & PoolFormer-S36~\cite{metaformer}    & 30.8      & 5.0G      & $224^{2}$ & 81.4  & CVPR'22   \\
        & ViTAEv2-S~\cite{vitaev2}            & 19.3      & 5.7G      & $224^{2}$ & 82.6  & IJCV'23\\
        & EATFormer-Small~\cite{eatformer}    & 24.3      & 4.3G      & $224^{2}$ & 83.1  & IJCV'24   \\
        & \ding{73}~EMOv1-20M~\cite{emo}        & 20.5      & 3.8G      & $224^{2}$ & 82.0  & ICCV'23   \\
        & \ding{72}~\textbf{EMOv2-20M}        & 20.1      & 4.0G      & $224^{2}$ & 83.3  & - \\
        \hline
        \multirow{6}{*}{\rotatebox{90}{\makecell[c]{\textbf{50M$\times$80M} \\ \textbf{Magnitude}}}}
        & ResNet-152~\cite{resnet,timm_resnet}& 60.1      & 11.5G      & $224^{2}$ & 82.0  & CVPR'16   \\
        & Swin-B~\cite{swinv1}                & 87.7      & 15.5G      & $224^{2}$ & 83.5  & ICCV'21   \\
        & PoolFormer-M48~\cite{metaformer}    & 73.4      & 11.6G      & $224^{2}$ & 82.5  & CVPR'22   \\
        & ViTAEv2-48M~\cite{vitaev2}          & 48.6      & 13.4G      & $224^{2}$ & 83.8  & IJCV'23\\
        & EATFormer-Base~\cite{eatformer}     & 49.0      & 8.9G      & $224^{2}$ & 83.9  & IJCV'24   \\
        & \ding{72}~\textbf{EMOv2-50M}        & 49.8      & 8.8G      & $224^{2}$ & 84.1  & - \\
        \toprule[0.12em]
        \end{tabular}
    }
\end{table}

\subsection{Visual Analysis between EMOv1/v2}

\begin{figure}[tp]
    \centering
    \includegraphics[width=1.0\linewidth]{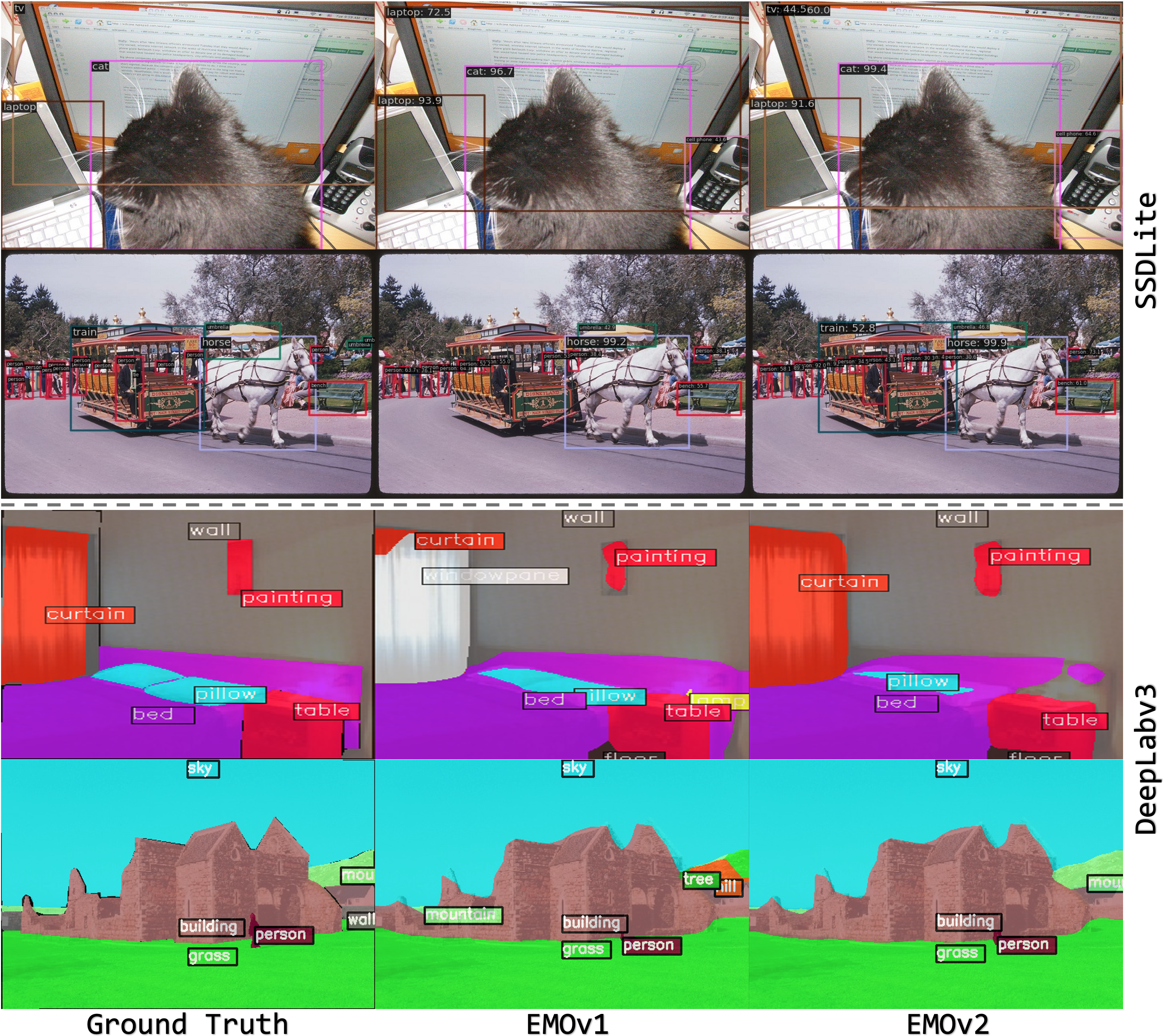}
    \caption{Qualitative comparisons between EMOv1/v2 on downstream SSDLite~\cite{mnetv3} and DeepLabv3~\cite{deeplabv3}. EMOv2 demonstrates higher accuracy in class and boundary detection. Zoom in for more details.}
    \label{fig:qualitative}
\end{figure}

\vspace{1mm} \noindent\textbf{Quantitative downstream visualization.} 
\cref{fig:qualitative}-Top presents the detection visualization results based on SSDLite. Compared to EMOv1, the improved EMOv2 demonstrates accurate classification and localization capabilities, even generalizing to objects that are missed in the ground truth. Thanks to the spanning attention mechanism, EMOv2 also achieves significant performance improvements in pixel-level dense prediction, as shown in \cref{fig:qualitative}-Bottom.

\begin{figure}[tp]
    \centering
    \includegraphics[width=1.0\linewidth]{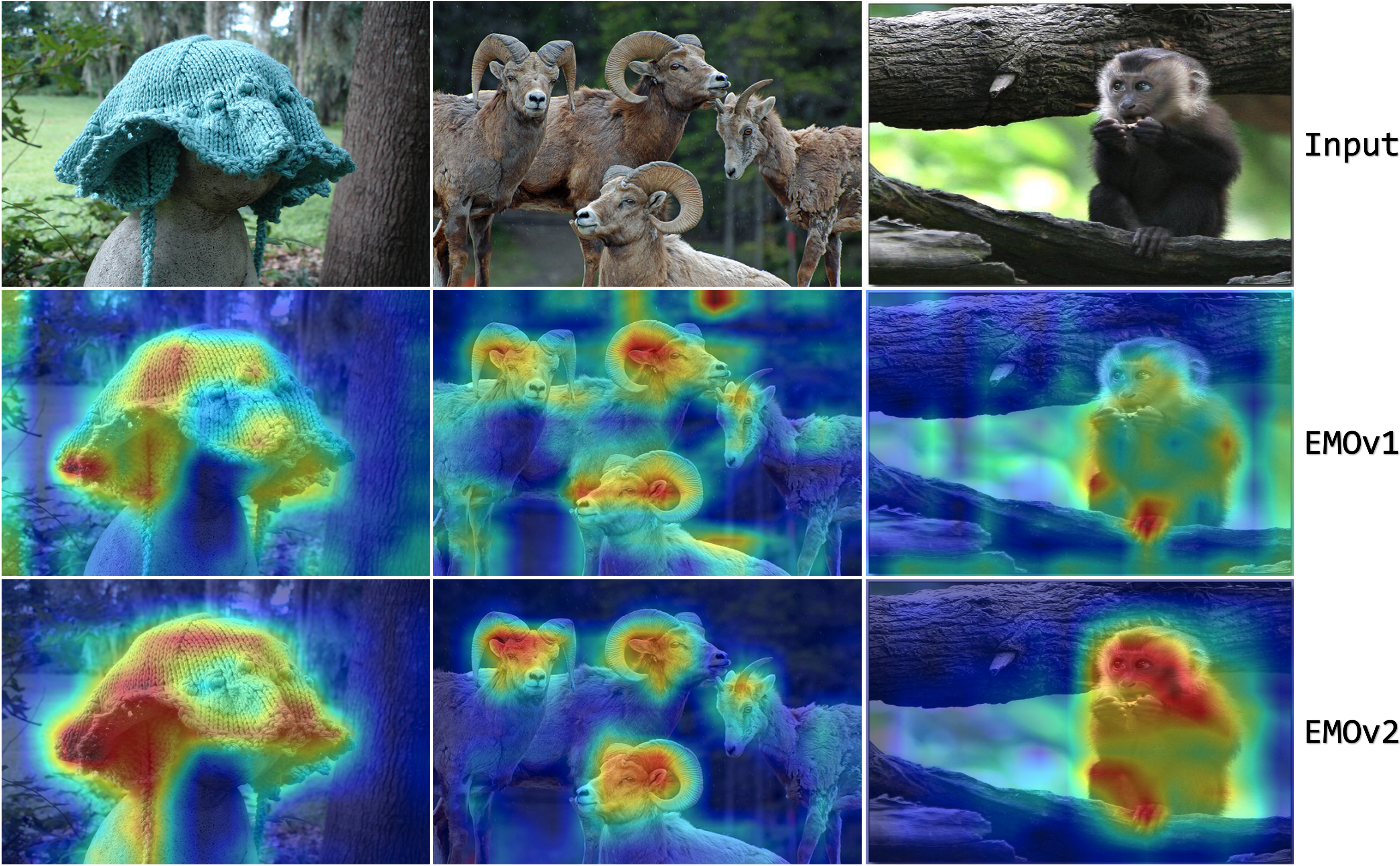}
    \caption{Visualizations by Grad-CAM. EMOv2 generates sharper and higher confidence attention maps than EMOv1.} 
    \label{fig:cam}
\end{figure}

\vspace{1mm} \noindent\textbf{Class activation mapping comparison.} 
\cref{fig:cam} presents the visualization results of Grad-CAM. The improved EMOv2 generates high-confidence class activations that are more closely aligned with the image subjects.

\subsection{Summary}

\begin{figure}[tp]
    \centering
    \includegraphics[width=1.0\linewidth]{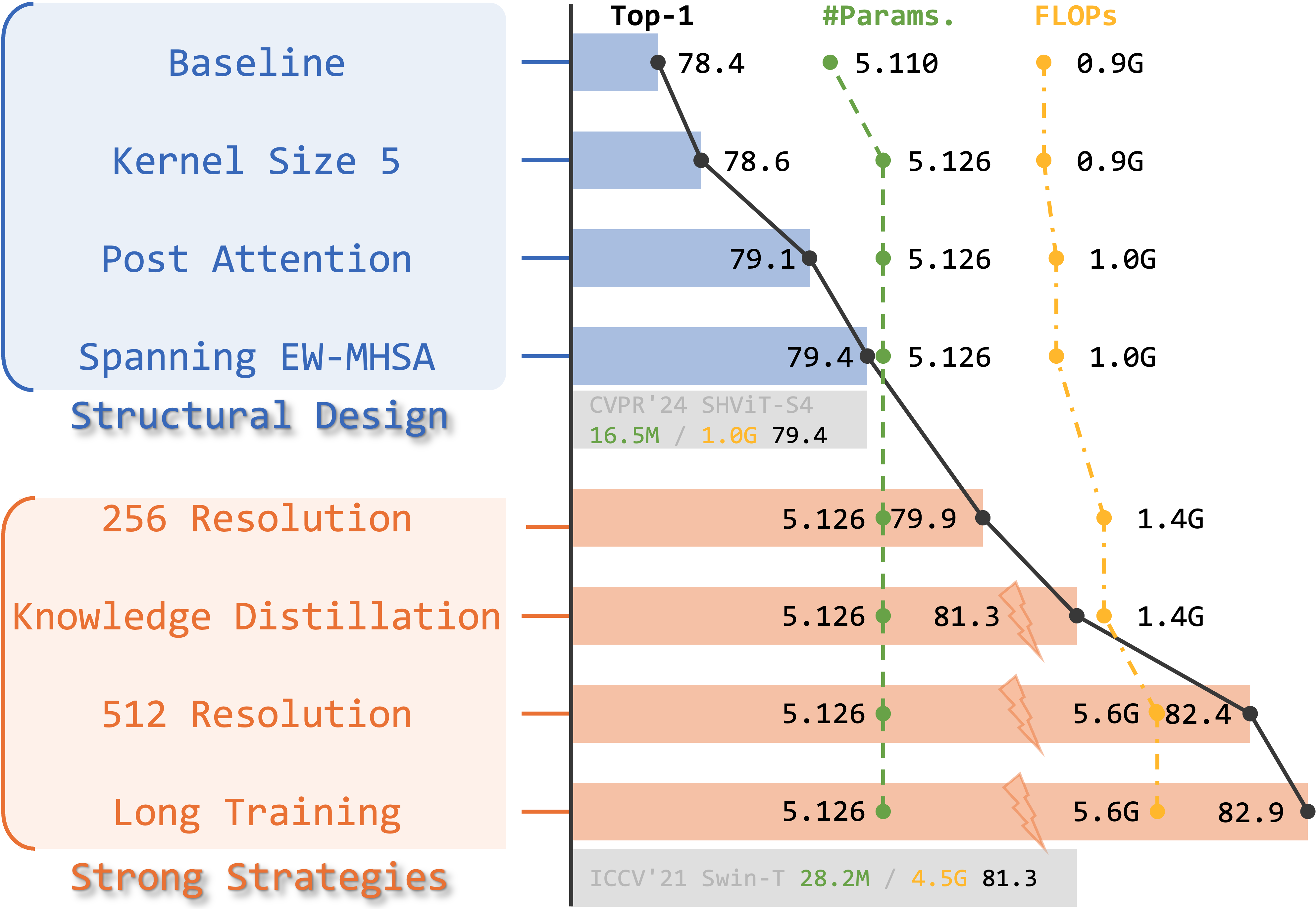}
    \caption{\textbf{Overall incremental trajectory from baseline to modern EMOv2 at the 5M magnitude.} Each line is based on a modification of the immediately preceding line. Detailed ablations in \cref{section:exp_ablation}. Parameters and FLOPs are marked in \textcolor{trajectory_green}{\textbf{green}} and \textcolor{trajectory_yellow}{\textbf{yellow}}.}
    \label{fig:ablation_trajectory}
\end{figure}

Starting from the EMOv1 baseline~\cite{emo}, we progressively explore factors influencing EMOv2 performance from the perspectives of \textit{structural design} and \textit{training strategy}. As shown in \cref{fig:ablation_trajectory}, the model parameters are controlled at 5.1M, and each structural improvement incrementally enhances the model's performance without additional parameter increase: 
\textbf{\textit{1)}} A larger kernel size improves the model's performance at the cost of only 0.016M parameters. 
\textbf{\textit{2)}} Post attention increases the Top-1 accuracy by 0.5 with an additional 0.1G FLOPs. 
\textbf{\textit{3)}} Spanning attention further enhances the model accuracy to 79.4, surpassing the baseline by +1.0$\uparrow$. Additionally, this operation significantly improves the performance of EMOv2 on downstream tasks, as shown in \cref{fig:v1v2}.
We use the structure at the end of the \textit{structural design} phase as our default EMOv2-5M, while higher resolution, extended training, and naive knowledge distillation strategies are employed to investigate the performance upper limits of our EMOv2 in the 5M parameter magnitude. The detailed structure can be viewed in the attached source code. 

\vspace{1mm} \noindent\textbf{Limitation discussion.} 
This study focuses on lightweight vision backbones and proposes EMOv2 model, extending them to the 20M and 50M parameter scales due to resource constraints. However, its Transformer-compatible architecture design potentially allows application to larger-scale vision backbones. Additionally, the spanning mechanism can be extended to the domain of large language models (LLMs), which warrants further exploration.

%% file: secs/5_conclusion.tex
\section{Conclusion} \label{section:con}
This work rethinks lightweight infrastructure from efficient IRB and effective components of Transformer in a unified perspective, proposing the abstracted concept of Meta Mobile Block for designing efficient models. 
Specifically, we deduce a modern infrastructural {\iirmb} to build a parameter-efficient attention-shared EMOv2, while extending it to dense prediction and generation fields by adapting {\iirmb} to different basic structures. Massive experiments on several downstream benchmarks demonstrate the superiority of our approach, and we also provide detailed studies and give some experimental findings on building an attention-based lightweight model.

%% file: secs/6_appendix.tex
\clearpage
\renewcommand\thefigure{A\arabic{figure}}
\renewcommand\thetable{A\arabic{table}}  
\renewcommand\theequation{A\arabic{equation}}
\setcounter{equation}{0}
\setcounter{table}{0}
\setcounter{figure}{0}
\appendix

\section*{Overview}
The supplementary material presents more comprehensive results of our EMOv2 to facilitate the comparison of subsequent methods:
\begin{itemize}
    \item \textbf{\cref{sec:supp_recipe}} provides detailed training recipes of various lightweight models trained on ImageNet-1K~\cite{imagenet} dataset. 
    \item \textbf{\cref{sec:supp_det}} provides more detailed object detection results using different frameworks on MS-COCO 2017~\cite{coco} dataset. 
    \item \textbf{\cref{sec:supp_seg}} provides more detailed semantic segmentation results using Mask R-CNN~\cite{maskrcnn} for multiple magnitudes of EMOv2 on ADE20K~\cite{ade20k} dataset. 
\end{itemize}

\begin{table*}[tp]
    \centering
    \caption{Comparison of \textbf{training recipes among popular and contemporary methods} and we employ the same setting in all experiments. Please zoom in for clearer comparisons. Abbreviations: MNet $\rightarrow$ MobileNet; MViT $\rightarrow$ MobileViT; EFormerv2 $\rightarrow$ EfficientFormerv2; GNet $\rightarrow$ GhostNet; NAS: Neural Architecture Search; KD: Knowledge Distillation; \#Repre.: Re-parameterization strategy.}
    \label{table:supp_train_recipe}
    \renewcommand{\arraystretch}{1.0}
    \setlength\tabcolsep{3.0pt}
    \resizebox{1.\linewidth}{!}{
        \begin{tabular}{cccccccccccccc}
        \toprule
        \multirow{2}{*}{Super-Params.} & \multirow{2}{*}{\makecell[c]{MNetv3~\cite{mnetv3}\\ \grayer{ICCV'19}}} & \multirow{2}{*}{\makecell[c]{ViT~\cite{vit}\\ \grayer{ICLR'21}}} & \multirow{2}{*}{\makecell[c]{DeiT~\cite{deit}\\ \grayer{ICML'21}}} & \multirow{2}{*}{\makecell[c]{MViTv1~\cite{mvitv1}\\ \grayer{ICLR'22}}} & \multirow{2}{*}{\makecell[c]{MViTv2~\cite{mvitv2}\\ \grayer{arXiv'22}}} & \multirow{2}{*}{\makecell[c]{EdgeNeXt~\cite{edgenext}\\ \grayer{arXiv'22}}} & \multirow{2}{*}{\makecell[c]{EFormerv2~\cite{efficientformerv2}\\ \grayer{ICCV'23}}} & \multirow{2}{*}{\makecell[c]{RepViT~\cite{repvit}\\ \grayer{CVPR'24}}} & \multirow{2}{*}{\makecell[c]{MogaNet~\cite{moganet}\\ \grayer{ICLR'24}}}   & \multirow{2}{*}{\makecell[c]{Vim~\cite{vmamba}\\ \grayer{ICLR'24}}} & \multirow{2}{*}{\makecell[c]{GNetv3~\cite{ghostnetv3}\\ \grayer{arXiv'2404}}} & \multirow{2}{*}{\makecell[c]{MNetv4~\cite{mnetv4}\\ \grayer{arXiv'2404}}} & \multirow{2}{*}{\makecell[c]{EMOv1/v2\\ \deepred{Ours}}} \\
        & & & & & & & \\
        \hline
        Epochs              & 300           & 300       & 300       & 300       & 300       & 300       & 300       & 300       & 300       & 300       & 600       & 500       & 300       \\
        Batch size          & 512           & 4096      & 1024      & 1024      & 1024      & 4096      & 1024      & 2048      & 1024      & 1024      & 2048      & 4096      & 2048      \\
        Optimizer           & RMSprop       & AdamW     & AdamW     & AdamW     & AdamW     & AdamW     & AdamW     & AdamW     & AdamW     & AdamW     & LAMB      & AdamW     & AdamW     \\
        Learning rate       & 6.4e$^{-2}$   & 3e$^{-3}$ & 1e$^{-3}$ & 2e$^{-3}$ & 2e$^{-3}$ & 6e$^{-3}$ & 1e$^{-3}$ & 4e$^{-3}$ & 1e$^{-3}$ & 1e$^{-3}$ & 5e$^{-3}$ & 4e$^{-3}$ & 6e$^{-3}$ \\
        Learning rate decay & 1e$^{-5}$     & 3e$^{-1}$ & 5e$^{-2}$ & 1e$^{-2}$ & 5e$^{-2}$ & 5e$^{-2}$ & 2.5$^{-2}$& 2.5$^{-2}$& 4$^{-2}$  & 1$^{-1}$  & 5$^{-2}$  & 1$^{-1}$  & 5e$^{-2}$ \\
        Warmup epochs       & 3             & 3.4       & 5         & 2.4       & 16        & 20        & 5         & 5         & 5         & 5         & 3         & 5         & 20        \\
        \hline
        Label smoothing     & 0.1           & \xmark    & 0.1       & 0.1       & 0.1       & 0.1       & 0.1       & 0.1       & 0.1       & 0.1       & 0.1       & 0.1       & 0.1       \\
        Drop out rate       & \xmark        & \cmark    & \xmark    & \cmark    & \xmark    & \xmark    & \xmark    & \xmark    & \xmark    & \xmark    & \cmark    & \cmark    & \xmark    \\
        Drop path rate      & \cmark        & \xmark    & \cmark    & \xmark    & \xmark    & \cmark    & \cmark    & \xmark    & \cmark    & \xmark    & \xmark    & \cmark    & 0.1       \\
        \hline
        RandAugment         & 9/0.5/1       & \xmark    & 9/0.5/1   & \xmark    & 9/0.5/1   & 9/0.5/1   & 9/0.5/1   & 9/0.5/1   & 7/0.5/1   & 9/0.5/1   & 9/0.5/1   & 15/0.7/2  & 9/0.5/1   \\
        Mixup alpha         & \xmark        & \xmark    & 0.8       & \xmark    & 0.8       & \xmark    & 0.8       & 0.8       & 0.1       & 0.8       & \xmark    & \xmark    & \xmark    \\
        Cutmix alpha        & \xmark        & \xmark    & 1.0       & \xmark    & 1.0       & \xmark    & 1.0       & 1.0       & 1.0       & 1.0       & \xmark    & \xmark    & \xmark    \\
        Erasing probability & 0.2           & \xmark    & 0.25      & \xmark    & 0.25      & \xmark    & 0.25      & 0.25      & 0.25      & 0.25      & \xmark    & -         & \xmark    \\
        Position embedding  & \xmark        & \cmark    & \cmark    & \xmark    & \xmark    & \cmark    & \cmark    & \cmark    & \cmark    & \cmark    & \xmark    & \xmark    & \xmark    \\
        Multi-scale sampler & \xmark        & \xmark    & \xmark    & \cmark    & \xmark    & \cmark    & \xmark    & \xmark    & \xmark    & \xmark    & \xmark    & \xmark    & \xmark    \\
        \hline
        NAS                 & \xmark        & \xmark    & \xmark    & \xmark    & \xmark    & \xmark    & \cmark    & \xmark    & \xmark    & \xmark    & \xmark    & \cmark    & \xmark    \\
        KD                  & \xmark        & \xmark    & \xmark    & \xmark    & \xmark    & \xmark    & \cmark    & \xmark    & \xmark    & \xmark    & \cmark    & \cmark    & \xmark    \\
        \#Repre.            & \xmark        & \xmark    & \xmark    & \xmark    & \xmark    & \xmark    & \xmark    & \cmark    & \xmark    & \xmark    & \cmark    & \xmark    & \xmark    \\
        \bottomrule
        \end{tabular}
    }
\end{table*}

\subsection{Detailed Training Recipes} \label{sec:supp_recipe}
Different SoTA lightweight methods~\cite{mnetv3,vit,deit,mvitv1,mvitv2,edgenext,efficientformerv2,repvit,moganet,vmamba,ghostnetv3,mnetv4} use various training recipes that could lead to potentially unfair comparisons, and we have summarized and compared these training strategies in \cref{table:supp_train_recipe}. Our training strategy is weaker, yet it achieves impressive results without employing strong training tricks. 

\subsection{Detailed Object Detection Results} \label{sec:supp_det}
Tab.~\ref{table:supp_det_ssdlite_retinanet} shows more detailed object detection results using SSDLite~\cite{mnetv3} and RetinaNet~\cite{retinanet} of our EMOv2 on MS-COCO 2017~\cite{coco} dataset, while Tab.~\ref{table:supp_det_maskrcnn} provide detailed object detection results using Mask R-CNN~\cite{maskrcnn}. 
    
\begin{table}[htp]
    \centering
    \caption{Detailed object detection performance using SSDLite~\cite{mnetv3} and RetinaNet~\cite{retinanet} of our EMOv2 on MS-COCO 2017~\cite{coco} dataset. $\dagger$: 512 $\times$ 512 resolution.}
    \label{table:supp_det_ssdlite_retinanet}
    \renewcommand{\arraystretch}{1.2}
    \setlength\tabcolsep{3.0pt}
    \resizebox{1.0\linewidth}{!}{
        \begin{tabular}{clcccccccc}
        \toprule[0.17em]
        & Backbone & \#Params $\downarrow$ & FLOPs $\downarrow$ & $mAP$  & $mAP^b_{50}$ & $mAP^b_{75}$ & $mAP^b_{S}$ & $mAP^b_{M}$ & $mAP^b_{L}$ \\
        \hline
        \multirow{8}{*}{\rotatebox{90}{\makecell[c]{SSDLite\\~\cite{mnetv3}}}} 
        & EMOv2-1M & 2.4 & 0.7G & 22.3 & 37.5 & 22.4 & 2.0 & 21.3 & 43.4 \\ 
        & EMOv2-1M$\dagger$ & 2.4 & 2.3G & 26.6 & 44.4 & 27.5 & 7.3 & 31.4 & 43.0 \\ 
        \cline{2-10}
        & EMOv2-2M & 3.3 & 1.2G & 26.0 & 43.0 & 26.5 & 3.6 & 26.6 & 50.2 \\ 
        & EMOv2-2M$\dagger$ & 3.3 & 4.0G & 30.7 & 49.8 & 31.7 & 9.9 & 37.1 & 47.3 \\ 
        \cline{2-10}
        & EMOv2-5M & 6.0 & 2.4G & 29.6 & 47.6 & 30.1 & 5.5 & 32.2 & 54.8 \\ 
        & EMOv2-5M$\dagger$ & 6.0 & 8.0G & 34.8 & 54.7 & 36.4 & 13.7 & 42.0 & 52.0 \\ 
        \cline{2-10}
        & EMOv2-20M & 21.2 & 9.1G & 33.1 & 51.9 & 33.9 & 8.9 & 36.8 & 57.3 \\ 
        & EMOv2-20M$\dagger$ & 21.2 & 30.3G & 38.3 & 58.4 & 40.7 & 17.9 & 45.2 & 54.6 \\ 
        \toprule[0.08em]
        \multirow{5}{*}{\rotatebox{90}{\makecell[c]{RetinaNet \\~\cite{retinanet}}}} 
        & EMOv2-1M & 10.5 & 142G & 36.9 & 57.1 & 39.0 & 22.1 & 39.8 & 49.5 \\ 
        & EMOv2-2M & 11.5 & 146G & 39.3 & 60.0 & 41.4 & 23.9 & 43.1 & 51.6 \\ 
        & EMOv2-5M & 14.4 & 158G & 41.5 & 62.7 & 44.1 & 25.7 & 45.5 & 55.5 \\ 
        & EMOv2-20M & 29.8 & 220G & 43.8 & 65.0 & 47.1 & 28.0 & 47.4 & 59.0 \\ 
        \bottomrule[0.12em]
        \end{tabular}
    }
\end{table}

\begin{table}[htp]
    \centering
    \caption{Detailed object detection performance using Mask RCNN~\cite{maskrcnn} of our EMOv2 on MS-COCO 2017~\cite{coco} dataset.}
    \label{table:supp_det_maskrcnn}
    \renewcommand{\arraystretch}{1.3}
    \setlength\tabcolsep{3.0pt}
    \resizebox{1.0\linewidth}{!}{
        \begin{tabular}{lcccccccc}
        \toprule[0.17em]
        \multirow{2}{*}{Backbone} & \multirow{2}{*}{\#Params $\downarrow$} & \multirow{2}{*}{FLOPs $\downarrow$} & $mAP$  & $mAP^b_{50}$ & $mAP^b_{75}$ & $mAP^b_{S}$ & $mAP^b_{M}$ & $mAP^b_{L}$ \\ 
        \cline{4-9}
        &  &  & $mAP$  & $mAP^m_{50}$ & $mAP^m_{75}$ & $mAP^m_{S}$ & $mAP^m_{M}$ & $mAP^m_{L}$ \\
        \hline
        \multirow{2}{*}{EMOv2-1M} & \multirow{2}{*}{21.2} & \multirow{2}{*}{165G} & 37.1 & 59.2 & 39.6 & 21.8 & 39.9 & 49.5 \\ 
        \cline{4-9}
        &  &  & 35.0 & 56.4 & 37.0 & 16.7 & 37.2 & 51.8 \\ 
        \hline
        \multirow{2}{*}{EMOv2-2M} & \multirow{2}{*}{22.1} & \multirow{2}{*}{170G} & 39.5 & 61.8 & 42.4 & 22.9 & 43.0 & 52.6 \\ 
        \cline{4-9}
        &  &  & 36.9 & 58.9 & 39.4 & 17.7 & 39.4 & 53.8 \\ 
        \hline
        \multirow{2}{*}{EMOv2-5M} & \multirow{2}{*}{24.8} & \multirow{2}{*}{181G} & 42.3 & 64.3 & 46.3 & 25.8 & 45.6 & 56.3 \\ 
        \cline{4-9}
        &  &  & 39.0 & 61.4 & 42.1 & 20.0 & 41.8 & 57.0 \\ 
        \hline
        \multirow{2}{*}{EMOv2-20M} & \multirow{2}{*}{39.8} & \multirow{2}{*}{244G} & 44.2 & 66.2 & 48.7 & 27.4 & 47.6 & 58.7 \\ 
        \cline{4-9}
        &  &  & 40.6 & 63.6 & 43.4 & 21.7 & 43.4 & 59.1 \\ 
        &  &  & 41.8 & 64.9 & 45.0 & 21.1 & 45.2 & 60.5 \\ 
        \toprule[0.12em]
        \end{tabular}
    }
\end{table}

\subsection{Detailed Semantic Segmentation Results} \label{sec:supp_seg}
Tab.~\ref{table:supp_seg} shows more detailed semantic segmentation results using DeepLabv3~\cite{deeplabv3}, Semantic FPN~\cite{semanticfpn}, SegFormer~\cite{segformer}, and PSPNet~\cite{pspnet} of our EMOv2 on ADE20K~\cite{ade20k} dataset, while Tab.~\ref{table:supp_seg_unet} provide detailed semantic segmentation results by adapting UNet with \iirmb.

\begin{table}[htp]
    \centering
    \caption{Detailed semantic segmentation performance using DeepLabv3~\cite{deeplabv3}, Semantic FPN~\cite{semanticfpn}, SegFormer~\cite{segformer}, and PSPNet~\cite{pspnet} to adequately evaluate our EMOv2 on ADE20K~\cite{ade20k} dataset.}
    \label{table:supp_seg}
    \renewcommand{\arraystretch}{1.2}
    \setlength\tabcolsep{6.0pt}
    \resizebox{1.0\linewidth}{!}{
        \begin{tabular}{clccccc}
        \toprule[0.17em]
        & Backbone & \#Params $\downarrow$ & FLOPs $\downarrow$ & mIoU  & aAcc & mAcc \\
        \hline
        \multirow{4}{*}{\rotatebox{90}{\makecell[c]{DeepLabv3}}} 
        & EMOv2-1M & 5.6 & 3.3G & 34.6 & 75.9 & 45.5 \\ 
        & EMOv2-2M & 6.6 & 5.0G & 36.8 & 77.1 & 48.6 \\ 
        & EMOv2-5M & 9.9 & 9.1G & 39.8 & 78.3 & 51.5 \\ 
        & EMOv2-20M & 26.0 & 31.6G & 43.3 & 79.6 & 56.0 \\ 
        \hline
        \multirow{4}{*}{\rotatebox{90}{\makecell[c]{FPN}}} 
        & EMOv2-1M & 5.3 & 23.4G & 37.1 & 78.2 & 47.6 \\ 
        & EMOv2-2M & 6.2 & 25.1G & 39.9 & 79.3 & 51.1 \\ 
        & EMOv2-5M & 8.9 & 29.1G & 42.3 & 80.8 & 53.4 \\ 
        & EMOv2-20M & 23.9 & 51.5G & 46.8 & 82.2 & 58.3 \\ 
        \hline
        \multirow{4}{*}{\rotatebox{90}{\makecell[c]{SegFormer}}} 
        & EMOv2-1M & 1.4 & 5.0G & 37.0 & 77.7 & 47.5 \\  
        & EMOv2-2M & 2.6 & 10.3G & 40.2 & 79.0 & 51.1 \\  
        & EMOv2-5M & 5.3 & 14.4G & 43.0 & 80.5 & 53.9 \\ 
        & EMOv2-20M & 20.4 & 36.8G & 47.3 & 82.1 & 58.7 \\ 
        \hline
        \multirow{4}{*}{\rotatebox{90}{\makecell[c]{PSPNet}}} 
        & EMOv2-1M & 4.2 & 2.9G & 33.6 & 75.8 & 44.8 \\ 
        & EMOv2-2M & 5.2 & 4.6G & 35.7 & 76.7 & 47.0 \\ 
        & EMOv2-5M & 8.1 & 8.6G & 39.1 & 78.2 & 51.0 \\ 
        & EMOv2-20M & 23.6 & 30.9G & 43.4 & 79.6 & 55.7 \\  
        \bottomrule[0.12em]
        \end{tabular}
    }
\end{table}

\begin{table}[htp]
    \centering
    \caption{Detailed semantic segmentation performance by adapting UNet with {\iirmb} on ADE20K~\cite{ade20k} dataset.}
    \label{table:supp_seg_unet}
    \renewcommand{\arraystretch}{1.2}
    \setlength\tabcolsep{6.0pt}
    \resizebox{1.0\linewidth}{!}{
        \begin{tabular}{lccccc}
        \toprule[0.17em]
        Backbone & \#Params $\downarrow$ & FLOPs $\downarrow$ & mIoU  & aAcc & mAcc \\
        \hline
        UNet-S5-D16 & 29.0 & 204G & 88.9 & 97.0 & 86.2 \\ 
        EMOv2-5M & 21.3 & 228G & 89.5 & 97.1 & 88.3 \\ 
        \bottomrule[0.12em]
        \end{tabular}
    }
\end{table}